\documentclass[acmsmall, nonacm]{acmart}

\AtBeginDocument{%
  }

\setcopyright{acmlicensed}
\copyrightyear{2025}

\usepackage[utf8]{inputenc} 
\usepackage[T1]{fontenc}    
\usepackage{amsfonts}       
\usepackage{hyperref}       
\usepackage{url}            
\usepackage{nicefrac}       
\usepackage{microtype}      
\usepackage{xcolor}         
\usepackage{tabularx} 

\usepackage[ruled,linesnumbered]{algorithm2e}
\usepackage{makecell, multirow}
\usepackage{float}
\usepackage{enumitem}
\usepackage{subcaption}
\usepackage{bm}
\usepackage{cleveref}
\usepackage{xspace}
\usepackage{soul} 

\newcommand{\project}{Centrum\xspace} 

\renewcommand{\vec}[1]{\bm{#1}}

\newcommand{\notcheckmark}{{$\surd$}\textsuperscript{\textcolor{black}{\kern-0.6em{\bf---}}}}

\newcommand{\sr}{SR\textsuperscript{2}\xspace}   
\newcommand{\nais}{NAIS\xspace}   
\newcommand{\tsen}{$\text{T}_{\text{sen}}$\xspace}   
\newcommand{\tto}[1]{$\text{T}_{#1\%}$\xspace}  

\newcommand{\envmysql}{MySQL8-SYSBENCH\xspace}
\newcommand{\envpg}{PG10-SYSBENCH\xspace}
\newcommand{\envsimsysbench}{MySQL5-SYSBENCH\xspace}
\newcommand{\envsimjob}{MySQL5-JOB\xspace}
\newcommand{\envsimtpcc}{MySQL5-TPCC\xspace}

\begin{document}

\title {Centrum: Model-based Database Auto-tuning with Minimal Distributional Assumptions}

\author{Yuanhao Lai}
\authornote{Both authors contributed equally to this research.}
\email{laiyuanhao@huawei.com}
\affiliation{%
	\institution{Huawei}
	\city{Shenzhen}
	\state{Guangdong}
	\country{China}
}

\author{Pengfei Zheng}
\authornotemark[1]
\authornote{Pengfei Zheng is the corresponding author.}
\email{zhengpengfei18@huawei.com}
\affiliation{%
	\institution{Huawei}
	\city{Shenzhen}
	\state{Guangdong}
	\country{China}
}

\author{Chenpeng Ji}
\email{jichenpeng@huawei.com}
\affiliation{%
	\institution{Huawei}
	\city{Shenzhen}
	\state{Guangdong}
	\country{China}
}

\author{Yan Li}
\email{liyan412@huawei.com}
\affiliation{%
	\institution{Huawei}
	\city{Shenzhen}
	\state{Guangdong}
	\country{China}
}

\author{Songhan Zhang}
\email{222010549@link.cuhk.edu.cn}
\affiliation{%
	\institution{The Chinese University of Hong Kong-Shenzhen}
	\city{Shenzhen}
	\state{Guangdong}
	\country{China}
}
\affiliation{%
	\institution{Huawei}
	\city{Shenzhen}
	\state{Guangdong}
	\country{China}
}

\author{Rutao Zhang}
\email{zhangrutao1@huawei.com}
\affiliation{%
	\institution{Huawei}
	\city{Shenzhen}
	\state{Guangdong}
	\country{China}
}

\author{Zhengang Wang}
\email{wangzhengang@huawei.com}
\affiliation{%
	\institution{Huawei}
	\city{Shenzhen}
	\state{Guangdong}
	\country{China}
}

\author{Yunfei Du}
\email{duyunfei5@huawei.com}
\affiliation{%
	\institution{Huawei}
	\city{Shenzhen}
	\state{Guangdong}
	\country{China}
}

\renewcommand{\shortauthors}{Yuanhao Lai, Pengfei Zheng et al.}

\begin{abstract}
Gaussian Process (GP)-based Bayesian optimization (BO), i.e., GP-BO, emerges as a prevailing model-based framework for DBMS (Database Management System) auto-tuning. However, recent work shows GP-BO-based DBMS auto-tuners are significantly outperformed by auto-tuners based on SMAC, which features random forest surrogate models; such results motivate us to rethink and investigate the limitations of GP-BO in auto-tuner design. We find that the fundamental assumptions of GP-BO are widely violated when modeling and optimizing DBMS performance, while tree-ensemble-BOs (e.g., SMAC) can avoid the assumption pitfalls and deliver improved tuning efficiency and effectiveness.  Moreover, we argue that existing tree-ensemble-BOs restrict further advancement in DBMS auto-tuning. First, existing tree-ensemble-BOs can only achieve distribution-free point estimates, but still impose unrealistic distributional assumptions on uncertainty (interval) estimates, which can compromise surrogate modeling and distort the acquisition function. Second, recent advances in (ensemble) gradient boosting, which can further enhance surrogate modeling against vanilla GP and random forest counterparts, have rarely been applied in optimizing DBMS auto-tuners.

To address these issues, we propose a novel model-based DBMS auto-tuner, \textbf{Centrum}. Centrum achieves and improves distribution-free point and interval estimation in surrogate modeling with a two-phase learning procedure of stochastic gradient boosting ensembles (SGBE). Moreover, Centrum adopts a generalized SGBE-estimated locally-adaptive conformal prediction to facilitate a distribution-free interval (uncertainty) estimation and acquisition function. To our knowledge, Centrum is the first auto-tuner that realizes distribution-freeness to stress and enhance BO's practicality in DBMS auto-tuning, and the first to seamlessly fuse gradient boosting ensembles and conformal inference in BO. Extensive physical and simulation experiments on two DBMSs and three workloads show that Centrum outperforms 21 state-of-the-art (SOTA) DBMS auto-tuners based on BO with GP, random forest, gradient boosting, OOB (Out-Of-Bag) conformal ensemble and other surrogates, as well as that based on reinforcement learning and genetic algorithms.
\end{abstract}


%


\maketitle

\section{Introduction}
\label{sec:intro}

Cloud data platforms serve largely diverse data analytics workloads over heterogeneous hardware. Such scale and complexity significantly challenge DBMS performance engineering, as correctly configuring DBMSs to adapt to varied workloads and hardware characteristics is a herculean task. Traditionally, DBMS tuning laboriously relies on DBA (Database Administrator)'s domain knowledge and trial-and-errors. Such human efforts can fail to scale and generalize for tremendous DBMS instances in the cloud. Moreover, modern DBMS is built with multiple hundreds of tunable knobs; the entangled interdependence between knobs and their combinatorially complex impact on DBMS performance severely challenge human reasoning and their final tuning efficacy. To address these problems, and motivated by the viral success of model-based optimization, i.e., Bayesian Optimization (BO) \cite{snoek2012practical, frazier2018tutorial} in real-world applications (e.g., AutoML), database practitioners \cite{duan2009tuning,van2017automatic,fekry2020tune,zhang2021restune,kanellis2022llamatune,Bin2022} have initiated a wave of building DBMS auto-tuners with BO-centric techniques.

The surrogate model plays a crucial role as it directly influences the effectiveness and efficiency of BO. BO fits a surrogate model to predict the mean (point estimate) and uncertainty (interval estimate) of DBMS performance under different configurations, then with the surrogate model, composes an acquisition function that, in each iteration of trials and errors, suggests the most promising configuration of the highest acquisition value. Acquisition function balances BO's exploitation and exploration and surrogates' point and interval estimation accuracy are decisive to BO's performance. Low point-estimate accuracy of the surrogate model incurs faulty exploitation that misguides the optimizer to erroneously enter regions with less-performant configurations. While low interval(uncertainty)-estimate accuracy incurs blind exploration with either under- or over-confidence.

However, existing DBMS auto-tuners, whether they use Gaussian Processes-surrogate-based BO (GP-BO) \cite{seeger2004gaussian} or tree-ensemble-surrogate-based (tree-ensemble-BO) \cite{lindauer2022smac3,hutter2011,PGBM2021}, still lack sufficient justification for their selected surrogate models. In this study, we find the accuracy of their surrogates can be severely undermined due to \textbf{misspecified model assumptions}, which fails to meet the intrinsic complex distributional characteristics of real DBMS performance measurements\footnote{This study focuses on the surrogate modeling component of existing database auto-tuning systems for DBMS performance. Other components such as knob selection and knowledge transfer, are important but out of the scope.}. We detail the limitations of existing DBMS auto-tuners as follows.

\begin{figure}[tb]
	\centering
	\begin{subfigure}[b]{0.24\textwidth}
		\captionsetup{belowskip=0pt}
		\centering
		\includegraphics[height=0.12\textheight,width=\textwidth, trim={0.2cm 0.15cm 0.25cm 0.2cm}, clip]{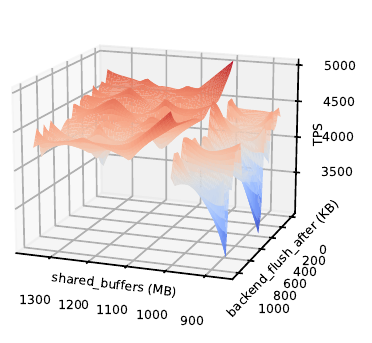}
		\caption{Non-Continuity}
		\label{fig:motivation_non_smooth0}
		\Description{The response surface between throughput measurements from PostgreSQL and different configurations, which shows their relationship is non-continuous and disjoint.}
	\end{subfigure}
	\hfill
	\begin{subfigure}[b]{0.24\textwidth}
		\centering
		\includegraphics[height=0.11\textheight,width=\textwidth]{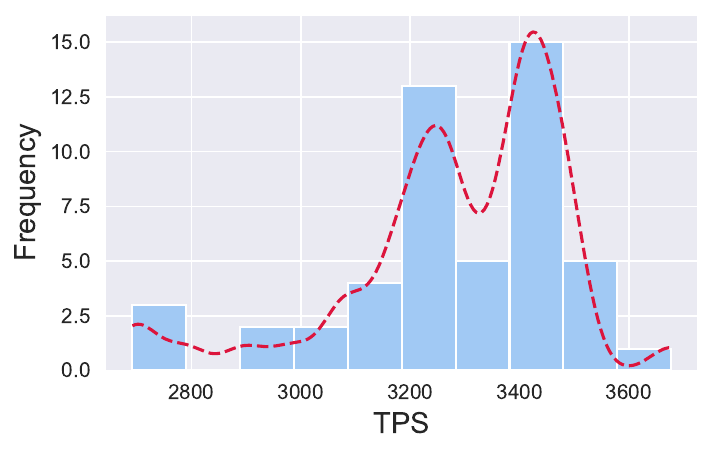}
		\caption{Non-Gaussianity}
		\label{fig:motivation_non_gaussian0}
		\Description{The histogram of throughput measurements from PostgreSQL at a given configuration under Sysbench workload, which shows the probability distribution of the performance metric is highly non-Gaussian.}
	\end{subfigure}
	\hfill
	\begin{subfigure}[b]{0.24\textwidth}
		\centering
		\includegraphics[height=0.11\textheight,width=\textwidth]{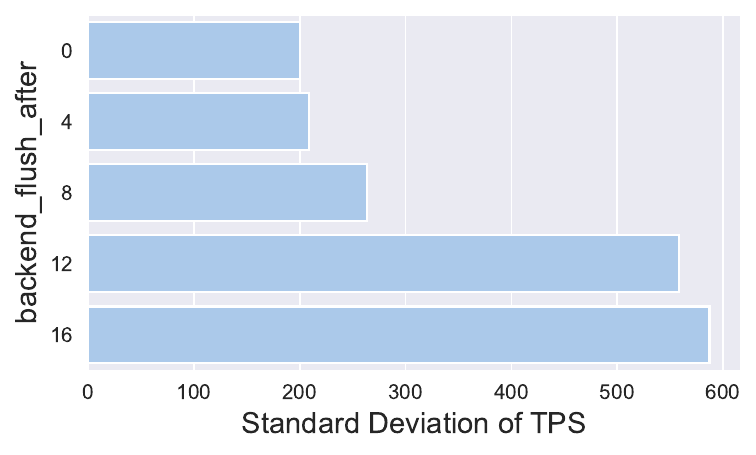}
		\caption{Heteroscedasticity}
		\label{fig:motivation_heteroscedasticity0}
		\Description{The bar chart of standard deviations for throughput measurements from PostgreSQL at different configuration settings under Sysbench Workload, which shows the performance metric is heteroscedastic.}
	\end{subfigure}
	\hfill
	\begin{subfigure}[b]{0.24\textwidth}
		\centering
		\includegraphics[height=0.11\textheight,width=\textwidth]{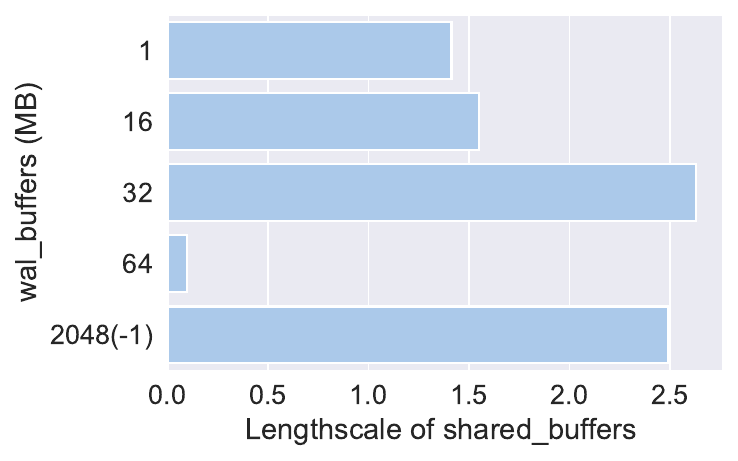}
		\caption{Non-stationarity}
		\label{fig:motivation_lengthscales0}
		\Description{The bar chart of the length scales regarding the parameter shared\_buffers, estimated by RBF kernel,  varied by the other parameter wal\_buffers, showing non-stationary characteristics.}
	\end{subfigure}
	\caption{
		Exemplify the violation of Gaussian Process's assumptions against PostgreSQL-v10.5-Sysbench auto-tuning
		- (a) non-smooth TPS (Transaction Per Second) surface over varying configurations; 
		- (b) empirical distribution of TPS is strikingly non-Gaussian and multi-modal;
		- (c) and (d)  heterogeneous noise levels and length scales. 
	}
	\label{fig:motivation_graphs0}
	\Description{Motivation examples to show the distributional characteristics of DBMS performance measurements are non-continuous, non-Gaussian, heteroscedastic, and non-stationary.}
\end{figure}

\begin{figure}[htb]
	\centering
	\begin{subfigure}[b]{0.24\textwidth}
		\centering	
		\includegraphics[height=0.17\textheight,width=\textwidth]{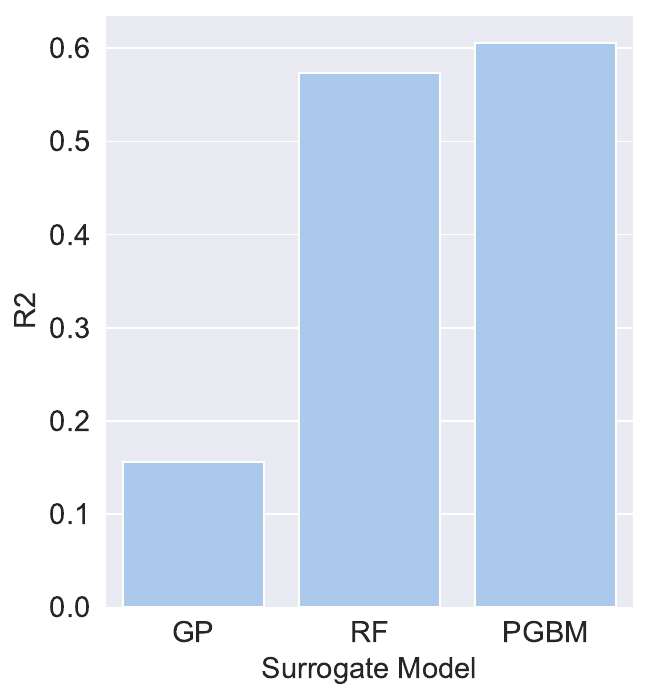}
		\caption{Point estimation}
		\label{fig:motivation_point_estimate}
		\Description{The bar chart of R-square with respect to different surrogate models including GP, RF and PGBM, which shows tree-based methods have much higher R-square than GP. Meanwhile, PGBM has higher R-square than RF.}
	\end{subfigure}
	\begin{subfigure}[b]{0.35\textwidth}
		\centering	
		\includegraphics[height=0.17\textheight,width=\textwidth]{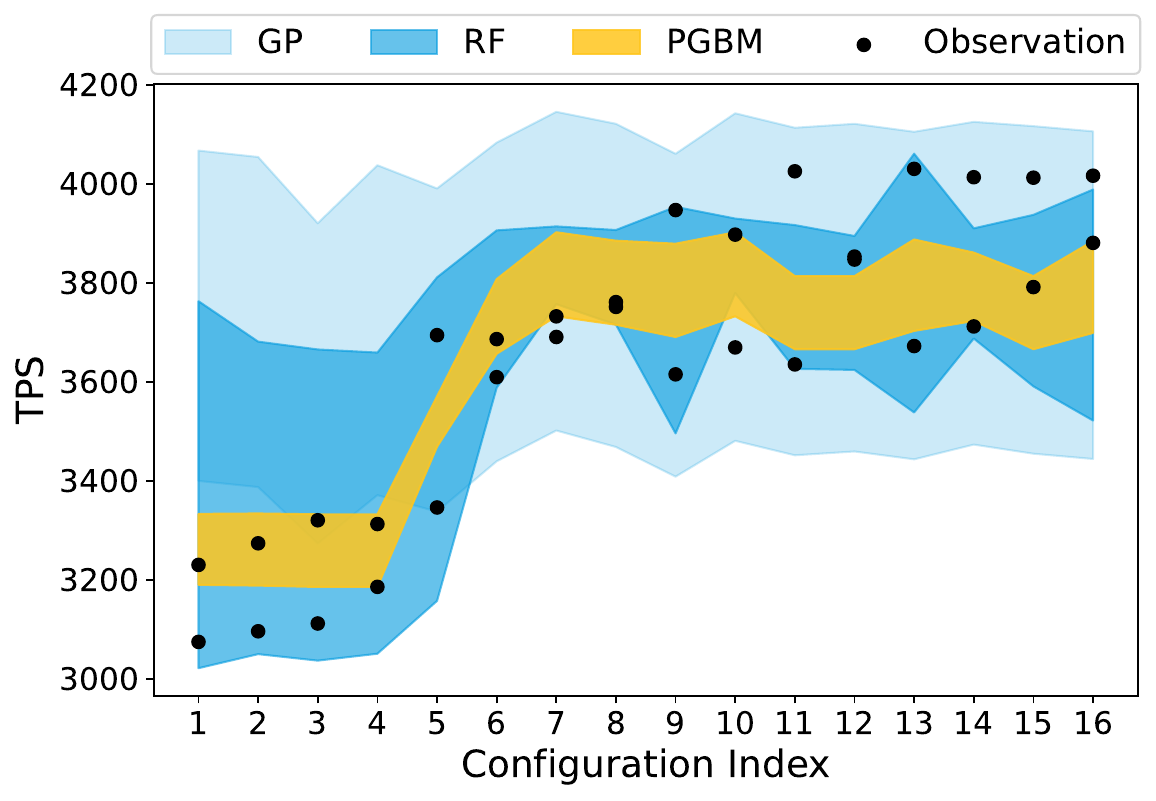}
		\caption{Interval estimation}
		\label{fig:motivation_interval_estimate}
		\Description{95\%-confidence interval Estimations of TPS along sixteen configuration inputs with respect to different surrogate models including GP, RF and PGBM, which shows GP's intervals are too wide, PGBM's intervals are too narrow, and RF's intervals fall between the two, but its coverage is still insufficient.}
	\end{subfigure}
	\begin{subfigure}[b]{0.35\textwidth}
		\centering	
		\includegraphics[height=0.17\textheight,width=\textwidth]{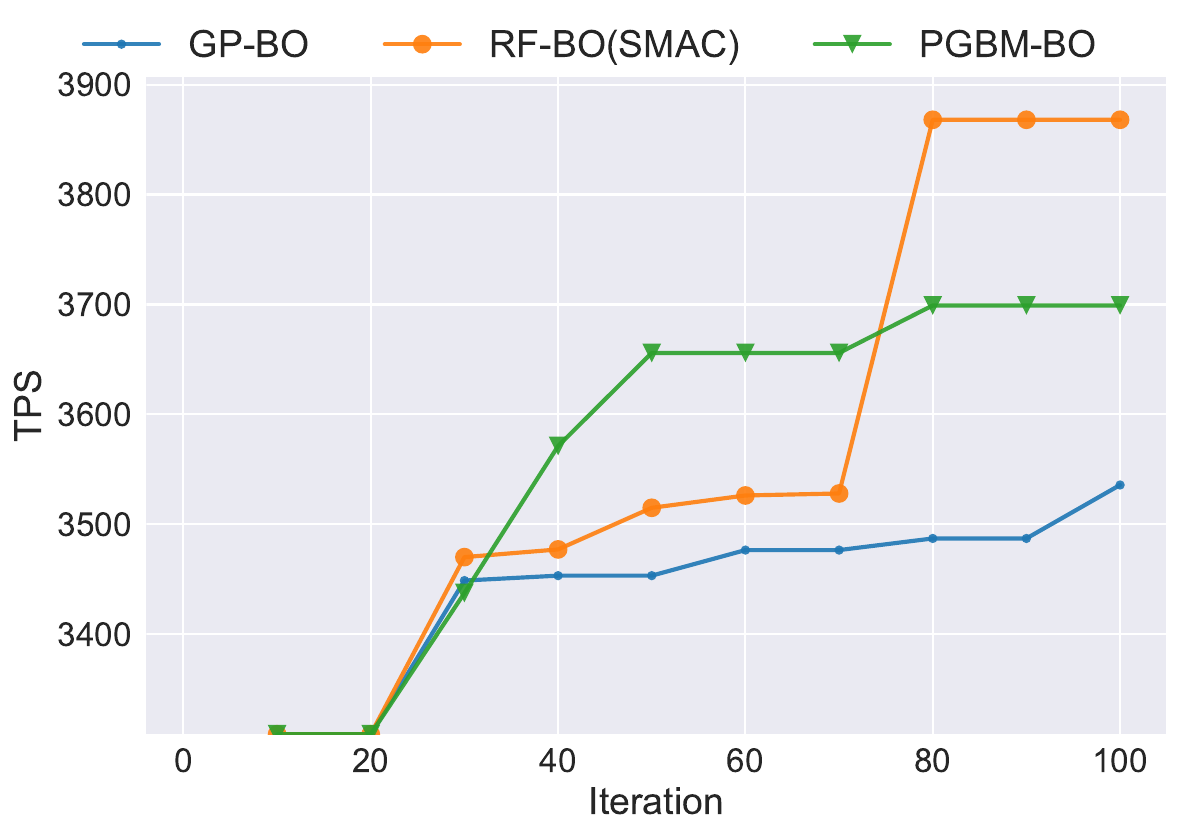}
		\caption{Tuning trajectory}
		\label{fig:motivation_tune_traj}
		\Description{The image is a line graph comparing the performance of three different optimization methods: GP-BO, RF-BO (SMAC), and PGBM-BO over 100 iterations. The graph indicates that RF-BO (SMAC) achieves the highest TPS, while PGBM-BO also performs well but stabilizes at a lower level than RF-BO (SMAC). GP-BO shows the most gradual improvement among the three.}
	\end{subfigure}
	\caption{
		Exemplify performance of surrogate model including GP, RF and PGBM in tuning PostgreSQL-v10.5-Sysbench
		- (a) $R^2$ of point estimations; 
		- (b) 95\%-confidence interval estimations;
		- (c) resulted BOs' tuning trajectories.}
	\label{fig:motivation_graphs_acc}
	\Description{}
\end{figure}

\textbf{Limitations of GP-BO-centric DBMS auto-tuners.}
GP-BO predominately prevails in DBMS auto-tuner design, which includes
iTuned \cite{duan2009tuning}, OtterTune \cite{van2017automatic}, Tuneful \cite{fekry2020tune}, ResTune \cite{zhang2021restune}, OnlineTune \cite{zhang2022towards}, ReLM \cite{kunjir2020black}, LlamaTune \cite{kanellis2022llamatune}, and LOCAT \cite{xin2022locat}.
However, \Cref{fig:motivation_non_smooth0,fig:motivation_non_gaussian0,fig:motivation_heteroscedasticity0,fig:motivation_lengthscales0} shows that, in real PostgreSQL (v10.5) tuning, \textbf{GP-BO's assumptions, (a) continuity (b) Gaussianaity (c) homoscedasticity and (d) stationarity fail to capture real DBMS performance characteristics that feature continuous-discrete-mixed, arbitrarily-distributed, heteroscedastic, non-stationary and noisy system measurements.} Such systematic assumption violation can invalidate GP's point and interval estimations and undermine the optimization effectiveness as stated by previous studies \cite{deshwal2021bayesian,wiebe2022robust, HEBO, snoek2014input}. As a result,
the $R^2$ of GP-surrogate's point-estimate is as low as $0.1$ (cf., \Cref{fig:motivation_point_estimate}; its poorly estimated interval is too loose (cf., \Cref{fig:motivation_interval_estimate}), to correctly navigate configuration exploration and severely distorts BO's acquisition function; finally, inaccurate surrogate modeling results in sub-optimal tuned DBMS performance (cf., \Cref{fig:motivation_tune_traj}).

\textbf{Limitations of tree-ensemble-BO in DBMS auto-tuning}.
Inherent discreteness and robustness of tree-ensemble models, including Random Forest (RF) \cite{breiman2001random} and Gradient Boosting Decision Trees (GBDT) \cite{friedman2001greedy}, make them well-suited for producing accurate point-estimates over high-dimensional, continuous-discrete-hybrid spaces filled with arbitrarily distributed, fluctuating DBMS measurements, as stated by prior research \cite{shahriari2015taking}. In particular, the RF-based BO, SMAC \cite{lindauer2022smac3},
has shown superior performance over multiple advanced GP-BOs in a prior DBMS benchmark study \cite{Bin2022}. \Cref{fig:motivation_point_estimate} shows that SMAC's RF surrogate shows over 40\% and 10\% improvements in point-estimate accuracy ($R2$) and tuned TPS compared with GP in PostgreSQL tuning.
Recently, SMAC begins to prevail in the recent design of DBMS auto-tuners  \cite{kanellis2022llamatune, pilotscope24}. However, we argue that limited tree-ensemble-BO frameworks other than SMAC exist, and existing tree-ensemble-BOs suffer from a few limitations to advance DBMS auto-tuning further. \textbf{(1) Existing tree-ensemble-BO that include SMAC and GBDT-based BO schemes (see below), only guarantee distribution-freeness for point estimates but not for interval estimates. Distributional assumptions (e.g., Gaussian) are still imposed in surrogate models' uncertainty quantification, which can distort tree-ensemble-BO's acquisition function.} Similar to GP, such assumption violation can result in less precise predicted intervals and adversely impact tuning effectiveness. \Cref{fig:motivation_interval_estimate}) shows that the RF surrogate of SMAC produces an over-loose (over-thin) interval for the left-sided  (right-sided) configuration region, which can cause overrated (underrated) exploration of poor-performance (optimal-performance) regions, and thereby, delays for missing finding the true optimum.
\textbf{(2) GBDT-based BO is promising but has rarely been applied in DBMS auto-tuning.} GBDT is widely acknowledged to have superior point-estimate predictive accuracy compared to RF in many real-world applications such as recommendation systems \cite{grinsztajn2022tree, chen2016xgboost,makridakis2022m5} but sheds limited spotlight on BO and DBMS tuning. 
This is arguably due to the fact that GBDTs' interval estimation \cite{malinin2021uncertainty} is difficult and existing schemes for GBDT uncertainty quantification are flawed. First, as aforementioned, existing interval estimation schemes for GBDTs, including NGBoost \cite{duan2020ngboost}, PGBM \cite{PGBM2021}, quantile regression \cite{skopt}, and virtual ensemble \cite{malinin2021uncertainty} all require unrealistic distributional assumptions (e.g., Gaussian). Second, NGBoost and PGBM estimate only data uncertainty while ignoring epistemic uncertainty \cite{gal2016uncertainty}. \Cref{fig:motivation_graphs_acc} shows that
PGBM, a SOTA GBDT model for both point and interval estimation, has 5.58\% higher $R^2$ compared to RF, but its estimated interval is over-thin has extremely poor coverage of true performance measurements. Finally, its tuned performance is even worse than that of SMAC.

\textbf{Limitations of OOB (Out-Of-Bag) conformal ensemble in BO and DBMS auto-tuning.} Recent advances of conformal inference \cite{xu2021conformal,shafer2008tutorial,angelopoulos2021gentle,lei2018distribution,colombo2023training} highlights a new direction to achieve complete interval estimation, with both data and epistemic uncertainty, for arbitrary learning algorithms. Conformal inference has demonstrated superiority for DBMS cardinality estimation \cite{thirumuruganathan2022prediction}, but has not been applied in BO and validated for DBMS auto-tuning. Existing work in the ML community combines conformal inference with RF, namely OOB (Out-Of-Bag) conformal ensemble \cite{linusson2020efficient}, which can make a natural extension and improvement over SMAC. However, we argue that such \textbf{straightforward OOB-extended BO has the following limitations and our experiments show OOB-extended SMAC can be outperformed by vanilla SMAC in real DBMS tuning}. (1) \textbf{OOB straightforwardly average out-of-bag learners \cite{linusson2020efficient,gupta2022nested}, which we find yields sub-optimal point and interval estimates}. (2) \textbf{ Locally adaptive conformal inference is difficult and existing ERC (Error Re-weighted Conformal) method can fail.} Vanilla conformal inference (e.g., split conformal) yields intervals with constant width. Error Re-weighted Conformal (ERC) techniques, employed in OOB \cite{papadopoulos2008normalized,lei2018distribution}, can make intervals to have locally adaptive (input-independent) width but recent work shows the instability issues of ERC due to sub-optimal difficulty estimation and conformity score transformation \cite{colombo2023training}. Moreover, current OOB is only for conformalized random forests \cite{johansson2014regression} but not for gradient boosting, which needs further design and validation in DBMS tuning.

To overcome these limitations, we propose \underline{\textbf{\project}}, a novel gradient-boosting-ensemble model-based optimization framework,
aiming at pushing DBMS auto-tuning effectiveness and efficiency to a new limit with optimized surrogate modeling. We summarize the core design and technical contributions of \project as below.
\begin{enumerate}
\item \project enhances the surrogates' point-estimate accuracy of DBMS auto-tuners with Stochastic Gradient Boosting Ensembles (SGBE) \cite{friedman2002stochastic, malinin2021uncertainty}. \project also adopts a second fine-tuning learning phase to produce the optimal ensemble of gradient-boosting machines, which can further boost point-estimate accuracy. \textbf{Physical and simulated experiments in DBMS auto-tuning show \project's two-phase learned SGBE surrogates show on average 9.5\% and 92.6\% higher point-estimate accuracy than other tree-ensemble-BO (e.g., RF and NGBoost) and GP-BO counterparts.} Note that \project's learns SGBE surrogates in a distribution-free manner.

\item \project employs an advanced conformal ensemble method to construct locally adaptive interval estimates for GBDTs. \textbf{\project's distribution-free conformalized intervals outperform existing auto-tuners' estimated interval by  27.6\% and 105.7\% w.r.t tightness and coverage, against tree- ensemble (e.g., RF and NGBoost) and GP-BO counterparts in physical and simulated DBMS tuning.} In addition, along with Monte Carlo integration of quantile functions, \project constructs acquisition function in a distribution-free manner, and to our knowledge, is the first practical model-based DBMS auto-tuner with minimal parametric and distributional assumptions.

\item \project improves OOB's straightforward average of out-of-bag base learners with an optimal ensemble that learns to minimize the weights of under-fitted and correlated base learners due to lack of data in DBMS-tuning, by a second fine-tuning procedure (mentioned above) that co-optimizes point and interval estimates with an elaborately designed score. Further, \project improves OOB's ERC conformal predictor with a SGBE-estimated, log-linear transformed difficulty measure. \textbf{Experiments show \project outperforms OOB-extended BO with 14.05\% higher tuned DBMS TPS.}

\item Overall, physical and simulated experiments show \textbf{\project exhibits 19.2\% and 29.0\% better tuned DBMS throughput or latency compared to 21 state-of-the-art (SOTA) DBMS auto-tuners} based on BO with GP, tree-ensemble, OOB-conformal ensemble and other surrogates, as well as that based on reinforcement learning and genetic algorithm.

\item Overall, \textbf{\project dominates in tuning efficiency, with a 4.2$\times$ speedup compared to existing methods.}
\end{enumerate}

\section{Preliminaries}
\label{sec:pre}

In this section, we formalize the DBMS tuning problem
as sequential optimization and discuss the keys of Bayesian optimization.

\subsection{DBMS Tuning as Sequential Optimization}

Existing DBMS auto-tuners, including BO with varied surrogate models, reinforcement learning, and the genetic algorithm, can be formalized as a sequential optimization procedure. 
At any intermediate iteration $t$, the auto-tuner requests the configuration optimizer to suggest a candidate configuration $\vec{x}_t$. Next, the database measures and collects performance feedback $y_t=f(\vec{x})$ (that is, throughput or latency) for $\vec{x}_t$. Then, by analyzing the collected trial-and-error observations, the configuration optimizer suggests a new promising configuration $\vec{x}_{t+1}$ for the next iteration. The suggestion and evaluation loop repeats until an iteration budget $T$ or a target DBMS throughput (or latency) is reached. 
Most DBMS auto-tuners \cite{duan2009tuning,van2017automatic,fekry2020tune,zhang2021restune,zhang2022towards,kunjir2020black,kanellis2022llamatune,xin2022locat,kanellis2022llamatune, pilotscope24} adopt BO as their configuration optimizers, 
which fit a surrogate model $\hat{f}(\vec{x})$ to evaluate the potential of contributions and suggest the most promising ones by maximizing an acquisition function.

\subsection{Dissecting Bayesian Optimization from Modeling Perspectives}

\textbf{BO Surrogate Model - point estimate and interval estimate}. 
The surrogate model $\hat{f}(\vec{x})$ is trained to predict a point estimate $\mu(\vec{x})$ and an interval estimate $(l(\vec{x}),u(\vec{x}))$ of the performance response $y$ under configuration $\vec{x}$. 
The point estimate $\mu(\vec{x})$ predicts the mean performance of a database under uncertain variations induced by workload fluctuations,  background interference (host OS and VM), epistemic (model) errors of $\hat{f}(\vec{x})$, etc. The interval estimate $(l(\vec{x}),u(\vec{x}))$, which is usually specified by the standard deviation $\sigma(\vec{x})$ or the quantiles of $\hat{f}(\vec{x})$, statistically quantifies the bounds of such variations (under a confidence level $\alpha$). 

\textbf{BO Acquisition Function - modeling exploitation and exploration with the surrogate model}.
The configuration optimizer, within each iteration, suggests a promising candidate configuration by maximizing the acquisition function. 
The acquisition function comprises and optimally trades off an exploitation part and an exploration part, which are modeled by the point and interval estimate of the surrogate model, respectively (cf. \Cref{fig:overview_bbo}).
Specifically, the exploitation part directs the optimizer to concentrate on candidate configurations speculated to be optimal, based on the current belief of $\mu(\vec{x})$ (i.e., predicted $\hat{f}(\vec{x})$ in expectation). The exploration part doubts the belief of $\mu(\vec{x})$ and favors a closer examination of the configurations with high uncertainty and unknown potentials, i.e., high variances  $\sigma^2(\vec{x})$. For instance, for the UCB (Upper Confidence Bound)\footnote{GP-UCB: Gaussian process optimization in the UCB bandit setting}\cite{srinivas2009gaussian} acquisition function $UCB(x)=\mu(\vec{x})$+$\sqrt{\beta}\sigma(\vec{x})$, $\mu(\vec{x})$ is the point-prediction-based exploitation and $\sqrt{\beta}\sigma(\vec{x})$ is the interval-prediction-based exploration ($\beta$ is a pre-specified constant); the configuration optimizer promotes configurations with maximal UCB values, i.e., ones with superior predicted DBMS performance and high (wide) predicted variances (intervals). Other acquisition functions EI (Expected Improvement) and PI (Probability of Improvement) adopt an analogous paradigm.

Therefore, both surrogate modeling and the acquisition function are essential for effective and efficient BO, and consequently, DBMS tuning. 
Here, surrogate modeling plays a more fundamental role in ensuring the quality of exploration and exploitation \cite{ament2023unexpected}.
As such, we focus on improving the surrogate modeling accuracy of both point and interval estimation to enhance model-based DBMS auto-tuners.
\section{Centrum: Conformalized ENsemble boosted TRee Uncertainty Model}
\label{sec:methodologies}

\begin{figure}[t!]
\centering
\includegraphics[width=0.75\linewidth]{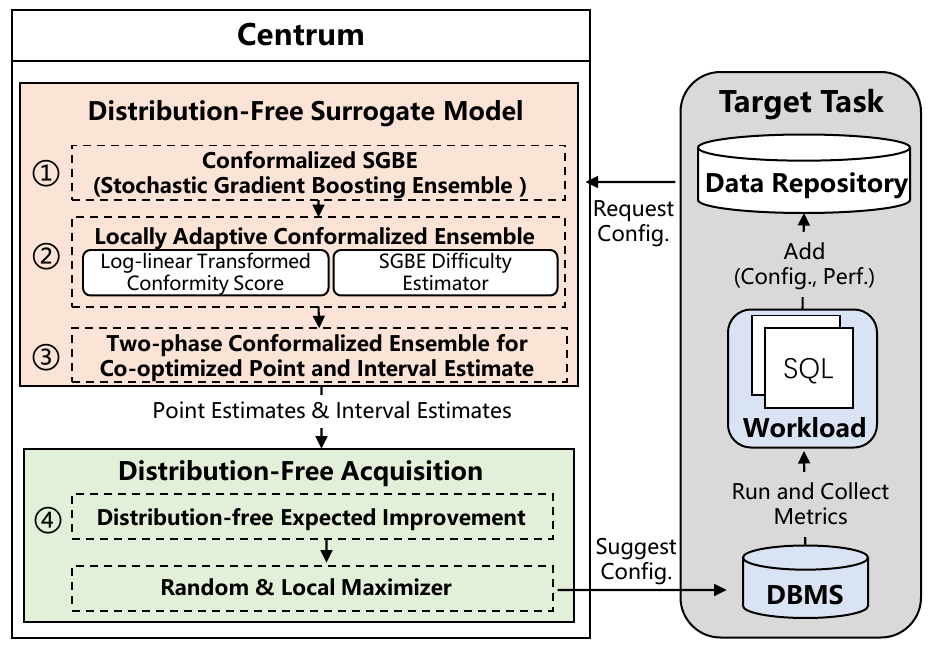}
\caption{Centrum framework overview}
\label{fig:overview_bbo}
\Description{The image shows a flowchart detailing our algorithm (Centrum) for optimizing database management systems (DBMS). It includes steps of fitting a distribution-free surrogate model and doing distribution-free acquisition inference, and running SQL workloads to collect performance metrics. The process integrates data repositories and DBMS to optimize configurations and improve performance.}
\end{figure}

As mentioned in \Cref{sec:intro}, 
the GP-BOs and tree-ensemble-BOs adopted by existing DBMS auto-tuners suffer from misspecified model assumptions that contradict the practical performance data,
which undermine the point- and interval-estimate accuracy of surrogate modeling and thus lower DBMS-tuning efficiency.
In this section, we resolve this limitation by presenting \project, 
a BO framework that seamlessly fuses GBDT and conformal ensembles to achieve an accurate distribution-free surrogate for both point and interval estimations.
\Cref{fig:overview_bbo} shows the overview of \project, which contains four steps.
We begin with the two most basic components in \Cref{sec: conformal_SGBE}, i.e.,
the distribution-free surrogate model of \textbf{conformalized stochastic gradient boosting ensemble (SGBE)} (Step 1) and a \textbf{distribution-free expected improvement acquisition computation} (Step 4) computed from an integration of quantile functions.
We then enhance the interval-estimate accuracy by \textbf{two locally adaptive conformalized ensemble methods} (Step 2) in \Cref{sec: local_conformal},
and enhance both the point- and interval-estimate accuracy via \textbf{a fine-tuning strategy to realize an optimal conformalized ensemble} (Step 3) in \Cref{sec: opti_conformal}.

\subsection{Distribution-Free BO with Conformalized Ensemble for Gradient Boosting}
\label{sec: conformal_SGBE}

By design, tree-ensemble surrogate models including Random Forest (RF) and Gradient Boosting Decision Trees (GBDT), can systematically avoid the assumption pitfalls of GP-BOs,
providing distribution-free point estimation for the non-continuous, non-Gaussian, heteroscedastic, and non-stationary relationship between configurations and DBMS performance metrics.
Compared to RF,
GBDT has demonstrated superior performance for tabular regression problems in numerous machine learning competitions and empirical studies \cite{grinsztajn2022tree, chen2016xgboost,makridakis2022m5}.
However, state-of-the-art tree-ensemble implementations, whether RF in SMAC \cite{lindauer2022smac3} or recent advances of probabilistic GBDTs such as NGB(Natural Gradient Boosting) \cite{duan2020ngboost}, PGBM (Probabilistic Gradient Boosting Machine)\cite{PGBM2021} and SGBE (Stochastic Gradient Boosting Ensembles) \cite{malinin2021uncertainty},
still assume the predictive distribution is Gaussian-distributed, and their interval estimation reduces to estimating the Gaussian variance. 
Meanwhile, no advanced GBDT-based BO has been applied to DBMS tuning yet. 
Directly applying GBDTs like NGB and PGBM to BO can cause under-exploration because they estimate only data uncertainty and omit epistemic uncertainty \cite{gal2016uncertainty}.
To fill this gap and achieve distribution-free interval estimation that incorporates both data and epistemic uncertainties in GBDT-based BO, we resort to recent advances such as Jackknife+ after bootstrap (J+aB) \cite{kim2020predictive} and conformal inference \cite{xu2021conformal}. 
We frame \project with conformal inference as it further lifts an unrealistic assumption (data exchangeability) made by J+aB \cite{xu2021conformal}. 
The resulting distribution-free GBDT-based BO consists mainly of the following two parts.

\subsubsection{Distribution-Free Conformalized Stochastic Gradient Boosting Ensemble Surrogate Model}
Conformal inference \cite{shafer2008tutorial,angelopoulos2021gentle,lei2018distribution} 
is appealing for uncertainty estimation due to its distribution-free properties, agnostic
to the model and data distribution with rigorous statistical guarantees of valid coverage in finite samples.
The general procedure for (split) conformal inference of a pre-trained regression model is simple. 
It firstly computes the conformity score $S(\vec{x},y)$ (e.g., typically an absolute prediction residual $S_i=S(\vec{x}_i,y_i)=|\mu(\vec{x_i})-y_i|$) on a calibration data set $D_{\text{cal}}=\{(x_i, y_i)\}_{i=1}^{m}$ for the pre-trained point-estimate regression model $\mu(\vec{x})$, and then uses the empirical quantiles of the conformity scores $\{S_i\}_{i=1}^m$ to quantify the distribution of the generation error for the pre-trained model. 
However, the calibration dataset $D_{\text{cal}}$ is required to be out-of-sample and hence reducing the size to train the regression model, which causes deterioration of point-estimate accuracy especially for small-sample scenarios such as DBMS-tuning.

To address the sample efficiency limitation of general conformal inference,
we make use of ensemble learning to construct \textbf{conformalized SGBE (Stochastic Gradient Boosting Ensemble)} as the surrogate model in \project. 
SGBE \cite{malinin2021uncertainty} is a strong gradient-boosting predictor that has superior point-estimate accuracy than a single GBDT, 
and the conformal ensemble method (i.e., EnbPI) \cite{xu2021conformal} offers a sample-efficient interval estimate (data and epistemic uncertainty estimation) in a distribution-free and model-free manner.

As illustrated in \Cref{alg:csgbe},
conformalized SGBE employs bagging \cite{breiman1996bagging} to form an average ensemble learner with each GBDT base learner trained on bootstrapped samples, where it further makes use of every out-of-bootstrap-sample data as the calibration set $D_{\text{cal}}$ to effectively compute the conformity scores of the corresponding base learner.
The our-of-sample data is aggregated to obtain conformity scores for the ensemble model, which is later used for constructing distribution interval estimations.

\subsubsection{Distribution-Free Acquisition Computation.}

Expected Improvement (EI) \cite{jones1998efficient,snoek2012practical} is arguably the most widely used acquisition function in BO.
It is designed to balance exploration and exploitation by utilizing surrogate predictions to estimate 
the expected improvement of a configuration over the best-observed performance that can be achieved by sampling a new point.
Formally, for a new configuration $\vec{x}$ and the best observed objective value $y^{*}$,
the EI acquisition function is given by $\text{EI}(\vec{x})=\mathbb{E}_{f(\vec{x})}[max(0,f(\vec{x})-y^*)]$.

However, for many black-box optimization tools of non-GP-based surrogates such as SMAC \cite{lindauer2022smac3} and open-Box \cite{JMLR:v25:23-0537}, 
the computation of EI still relies on Gaussian assumptions to get a simplified analytic formula based solely on the mean and variance estimations, which is violated in practice.
To avoid such a limitation,
we make use of interval estimates produced by the conformalized SGBE, and
propose a distribution-free method to compute the value of EI without relying on Gaussian assumptions.
We notice that an expectation can be derived in terms of the quantile function (See Chapter 3.2 of \cite{Random}),
and thereby, EI can be expressed as,
\begin{equation}
\centering
\text{EI}(\vec{x})=\int_{0}^{1}max(0,f(Q(\alpha,\vec{x})-y^{*})]\text{d}\alpha,
\label{eq:acq}
\end{equation}
where $Q(\alpha,\vec{x})$ is a $\alpha$-quantile function of $f(\vec{x})$.

Moreover, Theorem 1 of EnbPI \cite{xu2021conformal} shows that
the difference between $1-\alpha$ and the coverage for its $100(1-\alpha)\%$-confidence interval $[l_\alpha(\vec{x}), u_\alpha(\vec{x})]$ is bounded by
an error term that vanishes to zero as the sample size increases and the point estimate improves under regularity assumptions of stationary and strongly mixing errors,
and is approximately zero given high-quality point estimation.
By making a relaxation of symmetric intervals,
we can then approximate the quantile function $Q(\alpha,\vec{x})$ by $\hat{Q}(\alpha,\vec{x})=l_{2\alpha}(\vec{x})$ for $\alpha\in(0,0.5)$,
and $\hat{Q}(\alpha,\vec{x})=u_{2\alpha-1}(\vec{x})$ for $\alpha\in(0.5,1)$.
We then plug $\hat{Q}(\alpha,\vec{x})$ into \Cref{eq:acq} and apply Monte Carlo integration to get the value of EI without relying on Gaussian assumptions.
Finally, the random and local maximizer implemented in SMAC \cite{lindauer2022smac3} is used to maximize the derived EI to suggest a new configuration.

Overall, by seamlessly integrating conformal inference and bagging-based gradient boosting, we construct a distribution-free BO with the surrogate model of conformalized SGBE and a distribution-free EI acquisition computation, 
which achieves state-of-the-art accurate point estimation and distribution-free interval estimation that comprises both data and epistemic uncertainty. 

\begin{algorithm}[htb]
	\caption{(Generalized Locally adaptive) Conformalized SGBE}
	\label{alg:csgbe}
	\KwIn{
		Training Data (historic configuration-performance pairs up to $T$ iterations) $D=\{(\vec{x}_t, y_t)\}_{t=1}^{T}$,
		GBDT algorithm $\mathcal{A}$,
		indicator to control the local adaptation $I_{\text{local}}$,
		number of base learners $B$, 
		significance level $\alpha$, 
		and prediction input $\{\vec{x}_{\text{test}, i}\}_{i=1}^{n}$.
	}
	\KwOut{
		Mean estimations $\{\hat{h}^{\alpha}(\vec{x}_{\text{test}, i})\}_{i=1}^{n}$ and
		$100(1-\alpha)\%$ interval estimations  $\{C^{\alpha}(\vec{x}_{\text{test}, i})\}_{i=1}^{n}$
	}  
	\BlankLine
	\For{$b=1,\ldots,B$}{
		Randomly sample with replacement to obtain sub-data $D_{b}$ and its complementary data $\bar{X}_{b}$.
		Train a base model $M_{b} = \mathcal{A}(D_{b})$ .
	}	
	
	Initialize out-of-bag absolute calibration-error set  $\vec{\epsilon}_\text{oob}=\{\}$.
	
	\For{$t=1,\ldots,T$}{
		Let $\mathcal{B}_t=\{b\mid (\vec{x}_t, y_t) \notin D_{b}\}$.
		
		Compute $\hat{\epsilon}_{\text{oob},t} = |y_t - \hat{h}_{-t}(\vec{x}_t)|$ where 
		$\hat{h}_{-t}(\vec{x}_t) = \sum_{b\in\mathcal{B}_t}M_{b}(\vec{x}_t)/|\mathcal{B}_t|$.
		
		Update $\vec{\epsilon}_\text{oob}=\vec{\epsilon}\cup\{\hat{\epsilon}_{\text{oob},t}\}$.
	}	
	Let $D_{\epsilon,\text{oob}}=\bigl\{\bigl(\vec{x}_t, \log(\hat{\epsilon}_{\text{oob},t}\bigr)\bigr\}_{t=1}^{T}$ be the dataset for training auxilary model for predictive difficulty.
	
	\uIf{$I_{\text{local}}$ = ERC}{
		Let $\hat{g}(\vec{x}_t)=\exp(\mathcal{S}(\vec{x}_t))$,
		where an regression model $\mathcal{S} = \mathcal{A}\Bigl(D_{\epsilon,\text{oob}}\Bigr)$ is trained.
		
		Set conformality scores $S_t = \hat{\epsilon}_{\text{oob},t}/\hat{g}(\vec{x}_t)$, $\forall \hat{\epsilon}_{\text{oob},t} \in \vec{\epsilon}_\text{oob}$.
	}
	\uElseIf{$I_{\text{local}}$ = Generalized}{
		Compute $\hat{g}(\vec{x}_t)=\hat{g}^{\text{nested}}_{-t}(\vec{x}_t)$ according to \Cref{eq:nestconformal}.
		
		Set conformality scores $S_t = \log(\hat{\epsilon}_{\text{oob},t}) - \hat{g}(\vec{x}_t)$, $\forall \hat{\epsilon}_{\text{oob},t} \in \vec{\epsilon}_\text{oob}$.
	}
	
	Let $q^{\alpha} = (1-\alpha) \text{ quantile of } \{S_t\}_{t=1}^{T}$.
	
	\For{$i=1,\ldots,n$}{
		Let $\hat{h}(\vec{x}_{\text{test}, i}) = \sum_{t=1}^{T}\hat{h}_{-t}(\vec{x}_i)/T$ and $s_i=1$.
		
		\uIf{$I_{\text{local}}$ = Generalized}{
			Compute $C^{\alpha}(\vec{x}_\text{test})=[\hat{h}(\vec{x}_\text{test}) \pm  \exp(q^{\alpha}+g(\vec{x}_\text{test})]$,
		}
		\uElse{
			Let $s_i=\hat{g}(\vec{x}_{\text{test}, i})$ if $I_{\text{local}}$ = ERC.
			
			Compute $C^{\alpha}(\vec{x}_{\text{test}, i})=[\hat{h}^(\vec{x}_{\text{test}, i}) \pm s_i \cdot q^{\alpha} ]$.
		}
	}	
\end{algorithm}

\subsection{Locally Adaptive Conformalized Ensemble}
\label{sec: local_conformal}

Despite the fact that EnbPI estimates data and epistemic uncertainty in a distribution-free manner, there still exists gap when fusing it into the framework of BO. 
While EnbPI constructs a constant-width interval for its predictions,
constant interval width is problematic for BO. 
It reduces the uncertainty term in the acquisition function into a constant value and nullifies exploration. 
In addition, besides coverage, BO requires intervals to be tight, which can adapt to the right level of noise and uncertainties in different inputs. Constant interval width loses adaptive tightness. 
Thus, the key challenge is to produce locally-adaptive intervals for EnbPI, while not compromising the overall interval coverage.
By analyzing recent advances on locally adaptive conformal methods \cite{colombo2023training,johansson2014regression},  
we first introduce a general solution with Error Re-weighted Conformal (ERC) to mitigate this challenge,
and then propose a novel generalized locally adaptive method to further improve the interval efficiency (tightness) , as detailed below.

\subsubsection{Error Re-weighted Conformal}
A general method to allow for varying-width intervals is to apply the Error Re-weighted Conformal (ERC) technique \cite{papadopoulos2008normalized,lei2018distribution} in conformal inference.
We can use ERC to patch EnbPI to produce ERC-EnbPI, as shown in line 10 to 12 of \Cref{alg:csgbe}. 
Intuitively, ERC constructs an auxiliary predictor, besides the point-predictor, to estimate the residual error after point-prediction; then it shrinks (widens) the width of interval at a point $\vec{x}$ when its predicted residual at  $\vec{x}$ is low (high). The intuition is that the interval should be tight (wide) for a sample with low (high) predictive difficulty, which can be measured by the residual error. 
For instance, 
the out-of-bag conformalized (OOBC) random forests \cite{johansson2014regression} uses the random forest as the surrogate and additionally fits artificial neural networks (ANN) to predict the logarithm of the out-of-bag residual errors, which is used for normalizing the conformity score. 
Formally for conformalized SGBE, first, we use an auxiliary GBDT model $\hat{g}(\vec{x})$ to estimate the absolute residuals $\epsilon$ which is trained on the integrated out-of-bag residual dataset $D_{\epsilon,\text{oob}}=\bigl\{\bigl(\vec{x}_t, \log(\hat{\epsilon}_{\text{oob},t}\bigr)\bigr\}_{t=1}^{T}$, as presented in \Cref{alg:csgbe}. 
Second, we construct the ERC normalized conformity score  $S_{norm}(\vec{x}, y)=S(\vec{x}, y)/\hat{g}(\vec{x})$ and compute $q^{\alpha}$, i.e., the  $(1-\alpha) \text{ quantile of } \{S_{norm}(\vec{x}, y) | (\vec{x},y) \in D_\text{cal}\}$.
Third, ERC-EnbPI's estimated $100(1-\alpha)\%$ intervals becomes  $C^{\alpha}(\vec{x}_{\text{test}, i})=[\hat{h}(\vec{x}_{\text{test}, i}) \pm  \hat{g}(\vec{x}_{\text{test}, i}) \cdot q^{\alpha}]$,  and is locally-adaptive to $\vec{x}$.

\subsubsection{Generalized Locally Adaptive Conformalized Ensemble with Log-Linear Transformation and SBGE Estimator for Difficulty Measure}

A recent theoretical study \cite{colombo2023training} shows that ERC-based conformal methods can be further improved with a generalized transformed conformity score that guarantees marginal validity with input-dependent locally adaptive size.
In addition, we observe that existing ERC methods rely only on a single model to estimate the predictive difficulty.
This may lead to biased difficulty estimates in the OOB conformal scenarios \cite{johansson2014regression} because they 
train the auxiliary model on the out-of-bag calibration set
and apply it to predict the difficulty of samples that have been seen in the calibration dataset, which can result in under-estimated uncertainty.
This observation motivates us to propose a generalized locally adaptive conformal method with transformed conformity scores, and to use SGBE to enhance the difficulty estimator of existing OOB conformal methods, as stated below.

A generalized transformed conformity score for an observation $\vec{x}_\text{new}$ is defined as $\phi_{\vec{x}_\text{new}}(S|g)$, where $S=|\vec{x}-y|$ is the original absolute conformity score, and  $\phi_{\vec{x}_\text{new}}(S|g)$ is a differentiable monotonic function with respect to $S$
for arbitrary $\vec{x}_\text{new}$, where $g(\vec{x})$ is a learnable function.
To enhance ERC, 
we adopt the log-linear transformed conformity score, 
$\phi_{\vec{x}_\text{new}}(S|g)=\log(S)-g(\vec{x}_\text{new})$, which is empirically shown to 
have smaller interval size than ERC and the exponential transformed conformity score,
while maintaining the same or even better coverage.
The resulted $100(1-\alpha)\%$ intervals then becomes 
$C^{\alpha}(\vec{x}_\text{test})=[\hat{h}(\vec{x}_\text{test}) \pm  \phi_{\vec{x}_\text{test}}^{-1}(q^{\alpha}|g)]$, 
where
$\phi_{\vec{x}_\text{test}}^{-1}(q^{\alpha}|g)=\exp(q^{\alpha}+g(\vec{x}_\text{test}))$,
and $q^{\alpha}$ is the $(1-\alpha)$ quantile of the transformed out-of-bag conformity scores $\{\phi_{\vec{x}_\text{new}}(\hat{\epsilon}_{\text{oob},t})\}_{t=1}^{T}$.

Moreover, to avoid both training and inference of a single auxiliary model $g(\vec{x})$ on the same dataset for obtaining the predictive difficulty measures,
we can make use of SGBE again to get a "nested" out-of-bag estimator for the predictive difficulty of each sample. 
We first generate $B^{\text{nested}}$ bootstrapped subsets $D_{\epsilon,\text{oob},b}^{\text{nested}}$'s of 
the out-of-bag-residual dataset $D_{\epsilon,\text{oob}}$,
and train a base GBDT model $M_b^{nested}$ for each subset $D_{\epsilon,\text{oob},b}^{\text{nested}}$.
We then define the aggregated nested out-of-bag estimate for difficulty of $\vec{x}_t$ as,
\begin{equation}
\hat{g}^{\text{nested}}_{-t}(\vec{x}_t) = \sum_{b\in\mathcal{B}^{nested}_t} \frac{M_{b}^{nested}(\vec{x}_t)}{|\mathcal{B}^{nested}_t|},
\label{eq:nestconformal}
\end{equation}
where $\mathcal{B}^{nested}_t=\{b\mid (\vec{x}_t, y_t) \notin D^{nested}_{\epsilon,\text{oob},b}\}$.
$\hat{g}^{\text{nested}}_{-t}(\vec{x}_t)$ is then used to facilitate the generalized (log-linear) conformity scores as depicted in line 16 of \Cref{alg:csgbe}.

\subsection{Two-phase Conformalized Ensemble for Co-optimized Point and Interval Estimate}

\label{sec: opti_conformal}

DBMS tuning particularly emphasizes sample efficiency, which minimizes trial-and-error overheads and resource expenses on the users' side. Only the fly, provided a small set of configuration trials, vanilla bootstrapping (namely the 0.632 rule) in OOB \cite{linusson2020efficient, johansson2014regression} is likely to yield under-sampled datasets which results in insufficiently trained base GBDT learners. Second, as stated by previous studies \cite{zhou2012ensemble}, complementary and diverse base learners are fundamental in achieving an ensemble with good generalization. While, due to a lack of samples, bootstrapped base learners in OOB can not effectively guarantee such complementarity, and correlated, bootstrapped base learners can undermine the generalization of the final ensemble. We find that correlated base learners can be mitigated by using stochastic boosting (just as in SGBE) but cannot be effectively curbed. 

The issue of insufficiently trained and correlated base due to the lack of samples in the vanilla OOB method can degrade the point and interval estimate of auto-tuners' surrogate models. Therefore, we propose a second conformal ensemble fine-tuning phase that co-optimizes point and interval estimation accuracy.

We first define two optimization objective metrics \textbf{\sr} and \textbf{\nais} to respectively quantify the point-estimate accuracy and the interval-estimate accuracy of the surrogate model.
Subsequently, we propose a two-phase training methodology for SGBE,
where the first phase is to train conformalized SGBE as detailed in \Cref{sec: local_conformal},
and the second phase is a fine-tuning process that optimizes both \textbf{\sr} and \textbf{\nais} to achieve optimal ensemble.
A similar concept can be found in a seminar paper called GASEN \cite{GA2005},
which uses a genetic algorithm to select an optimal subset of individual networks to form a neural network ensemble
that optimize the point-estimate accuracy (i.e., minimal mean-squared-error).
Our approach differs from theirs in two significant ways. 
Firstly, our approach aims to improve both point and interval estimations instead of only point estimation.
Second, we not only adjust the ensemble composition but also the parameters of base models (optimal weights and truncation) to form optimal conformalized SGBE, which best trades off between the individual model performance and diversity.

\subsubsection{Accuracy Metrics of Surrogates.}
Given the configurations $\{\vec{x}_t\}\}_{t=1}^n$
the performance metrics $\vec{y}=\{\vec{x}_t\}\}_{t=1}^n$,
the surrogate's mean predictions $\vec{\mu} = \{\mu(\vec{x}_t)\}_{t=1}^{T}$,
and $100(1-\alpha)\%$-confidence interval predictions
$(\vec{l}_\alpha, \vec{u}_\alpha) = (\{l_\alpha(\vec{x}_t) \}_{t=1}^{T}, \{u_\alpha(\vec{x}_t)\}_{t=1}^{T})$,
the quality measurements for surrogate models are defined as follow, 
which will be used to facilitate the two-phase training of SGBE and the later experimental evaluation in \Cref{sec:evalution}.

(a) \textbf{\sr (Surrogate Coefficient of Determination)} measures the point-estimate accuracy with the coefficient of determination, 
$\text{\sr}(\vec{y},\vec{\mu})$ = $1 - \sum_{t=1}^{T}(y_t - \mu(\vec{x}_t))^2$$/ \sum_{t=1}^{T} (y_t - \bar{y})^2$. 
A higher \sr entails more accurate point estimations of the surrogate model and more valid exploitation.

(b) \textbf{\nais (Normalized Aggregate Interval Score)} measures the coverage and tightness jointly of the surrogate model's interval estimations over a series of confidence levels
of $\{\alpha_k\}_{k=1}^K$'s ranging from 0.01 to 0.99.
We first define the non-normalized metric AIS as,
$$\text{AIS}(\vec{y},\vec{l},\vec{u}) = \sum_{t=1}^{T} \sum_{k=1}^{K} \alpha_k \text{IS}_{\alpha_k}(y_t, l_{\alpha_k}(\vec{x}_t), u_{\alpha_k}(\vec{x}_t))/n,$$
where the metric IS (Interval Score), $\text{IS}_{\alpha}(y, l, u) = (u-l) + 2[(l-y)_{+}+(y-u)_{+}]/\alpha$,
evaluates a specific confidence level $\alpha$;
$(y)_{+}=\text{max}(0,y)$ is a hinge function; the left term $u-l$ quantifies the interval width and 
the right term $(l-y)_{+}+(y-u)_{+}$ is a loss to penalize when the true performance $y$ is either above the upper bound $u$ or below the lower bound $l$.  
\nais is then defined as 
$\text{\nais}(\vec{y},\vec{l},\vec{u})=(\text{AIS}_{\text{base}}-\text{AIS}(\vec{y},\vec{l},\vec{u}))/\text{AIS}_{\text{base}}$,
where $\text{AIS}_{\text{base}}$ is the AIS score for a $100(1-\alpha)\%$ interval predictor generated
by a Gaussian distribution with the mean and standard deviation equal to their empirical statistics of observed performance values.
A higher \nais indicates good interval tightness and coverage for all confidence levels.

\subsubsection{Fine-tuning as Constrained Optimization.}
\label{sec:finetune}
The proposed fine-tuning process aims to optimize both \sr and \nais of the trained conformalized SGBE from \Cref{alg:csgbe}
by specifying the ensemble model weights $\vec{w}=\{w_b\}_{b=1}^{B}$ 
and the trimmed rates $\vec{\lambda}=\{\lambda_b\}_{b=1}^{B}$ for each based model.
Formally, 
$M_{b}(\cdot|\lambda_b)$ denote the model obtained by keeping only the first $\lambda_b\in(0,1]$ percentage of decision trees from the base model $M_{b}$, 
which is trained on sub-data $D_{b}$ as shown by line 2 of \Cref{alg:csgbe},
while $w_b$ denote the weights associated with the base model $M_{b}$
in constructing a weighted ensemble model 
$\tilde{h}_{-t}(\vec{x}_t|\vec{w},\vec{\lambda}) = \sum_{b\in\mathcal{B}_t} w_b M_{b}(\vec{x}_t|\lambda_b)/\sum_{b\in\mathcal{B}_t} w_b$ 
to obtain an out-of-bag prediction for sample $\vec{x}_t$
where $\mathcal{B}_t=\{b\mid (\vec{x}_t, y_t) \notin D_{b}\}$.

To avoid the difficulty in solving the Pareto Front for bi-objective optimization of \sr and \nais,
we formulate the proposed fine-tuning process as 
solving the following constrained optimization of a conjugate measure, WIS (Weighted Interval Score),
evaluated via out-of-bag samples, i.e.,
\begin{align*} 
	\underset{\vec{w},\vec{\lambda}}{\text{max}} &\phantom{11} 
	\text{WIS}(\vec{w}, \vec{\lambda}) = \text{\sr}\bigl(\vec{y},\vec{\mu}_{oob}(\vec{w},\vec{\lambda})\bigr) + \\
	&\phantom{111 \text{WIS}(\vec{w}, \vec{\lambda}) = }
	\text{\nais}\bigl(\vec{y},\vec{l}_{oob}(\vec{w},\vec{\lambda}),\vec{u}_{oob}(\vec{w},\vec{\lambda})\bigr) \\
	\text{s.t.} 
	&\phantom{11}  w_b \in \{0,1\}, b=1,\ldots,B, \\
	&\phantom{11}  0\leq \lambda_b \leq 1, b=1,\ldots,B, \\
	&\phantom{11}  \sum_{b\in\mathcal{B}_t} w_b > 0, t=1,\ldots,T,
\end{align*}
where the out-of-bag point and interval estimates are
$\vec{\mu}_{oob}(\vec{w},\vec{\lambda})=\{\tilde{h}_{-t}(\vec{x}_t|\vec{w},\vec{\lambda})\}_{t=1}^{T}$,
$\vec{l}_{oob}(\vec{w},\vec{\lambda})=\{\tilde{h}_{-t}(\vec{x}_t|\vec{w},\vec{\lambda})-\hat{g}(\vec{x}_{t}) \cdot q_{i}^{\alpha_k}\}_{t=1,k=1}^{T,K}$,
and
$\vec{u}_{oob}(\vec{w},\vec{\lambda})=\{\tilde{h}_{-t}(\vec{x}_t|\vec{w},\vec{\lambda})+\hat{g}(\vec{x}_{t}) \cdot q_{t}^{\alpha_k}\}_{t=1,k=1}^{T,K}$.
The last constraint $\sum_{b\in\mathcal{B}_t} w_b > 0$ for $t=0,\ldots,T$ ensure that
every sample $x_t$ will have an out-of-bag estimator $\tilde{h}_{-t}(\cdot)$,
and thus maintains sample efficiency for training the surrogate model.

Finally, We employs the cross-entropy method (CEM) \cite{rubinstein2004cross},
a Monte Carlo method that enables efficient combinatorial and continuous problem-solving.
Once the solution $\vec{\hat{w}}$ and $\vec{\hat{\lambda}}$ are found,
we replace the out-of-bag predictor $\hat{h}_{-t}$ in line 7 and 18 of \Cref{alg:csgbe} 
by $\tilde{h}_{-t}(\vec{x}_t|\vec{\hat{w}},\vec{\hat{\lambda}}$,
and update the conformal scores $\vec{\epsilon}$ and the locally adaptive estimator $\hat{g}(\cdot)$ accordingly.

\section{Evaluation}\label{sec:evalution}

\subsection{Experiment methodology}
\label{sec:experiment_method}

\textbf{Diverse experiment setups and comprehensive evaluation.} We set up diverse experiment settings to evaluate \project and the baseline optimizes; each set corresponds to a unique combination of DBMS, query workload, and configuration-parameter set. (cf., \Cref{tab:eval_task}). Our experiments collect results over 462 runs of DBMS auto-tuning procedures for 21 state-of-the-art model-based optimizers plus \project, on three DBMS (i.e., MySQL-v8.0, MySQLv-5.7, and PostgreSQL-v10.5), three OLTP and OLAP workloads (JOB, TPCC, and Sysbench), \textbf{which in total account for 1068 VM-hours (or 44.5 VM-days).}

\textbf{Assuring evaluation validity.} (a) To avoid systematic, inherent evaluation bias that possibly exists in our system setting, we adopt an open benchmark of DBMS auto-tuning to reinforce evaluation objectivity. The benchmark contains ML-based (Random Forest) DBMS-simulators which output the performance response of the fitted DBMS for any input configuration. The simulators are trained with real MySQL-v5.7 measurements on VMs with  8 vCPUs and 16GB RAM. The other benefit of using a simulator is to keep experimentation economic; simulator runs cut off 66\% of cloud VM-hours. The benchmark also releases the datasets used to train the RF models (simulators). To remove evaluation bias, we re-train simulators on the datasets with transformers. (See later explanation.) (b) Moreover, we set three (five) independent executions of each physical (simulator) experiment and reveal not only the average result of each optimizer but also the variation of results across executions. (c) Each execution of an optimizer's tuning procedure consists of 100 iterations. For BO optimizers, the first 20 iterations are generated randomly using a Sobol sampler, to train their surrogate models. To eliminate unfair comparison of optimizers' learnability and optimizability due to the random quality of initial samples, we draw the Sobol samples in advance and feed them to individual BO optimizers.

\textbf{Removing DBMS-simulator bias in open benchmark with tabular transformer.} 
The ML-based DBMS simulator in the open benchmark \cite{Bin2022} are random forest model. By virtue of model-structure homogeneity, tree-ensemble optimizers can have falsely boosted optimizability over other schemes such as GP-BOs, as their surrogate models can efficiently learn to replicate a tree-ensemble-structured simulator. To remove the structural bias of tree-structured DBMS-simulators, we re-fit the datasets released by the open benchmark and replace the RF-base simulators with transformer-based simulators \cite{mcelfresh2024neural, borisov2022deep}. We consider three representative transformer models for tabular data regression (tabular transformers) including SAINT \cite{somepalli2021saint}, FT-Transformer (FT) \cite{gorishniy2021revisiting}, and AutoInt \cite{song2019autoint}, as well as other DNN models including multilayer perceptron (MLP) and ResNet \cite{gorishniy2021revisiting}. We compare the accuracy (under 80/20-split holdout validation) of different simulators in \Cref{tab:nnsim}. SAINT achieves the highest simulation accuracy among all DNN simulators. Thus, we re-train simulators with the SAINT tabular transformer model, which uses column-attention, i.e., attention between features, and row-attention, i.e., attention between samples, to extrapolate and simulate database performance responses. Besides, we argue that the transformer is known to have extremely high structure-complexity, which renders a fair representation of hardness to all BO surrogate models.

\setlength{\tabcolsep}{2pt} 
\begin{table}[t]
	\centering
	\caption{Experiment setup}
	\label{tab:eval_task}
	\begin{tabularx}{\linewidth}{
			>{\raggedright\arraybackslash}>{\hsize=1.2\hsize}X
			>{\raggedright\arraybackslash}>{\hsize=0.3\hsize}X
			>{\centering\arraybackslash}>{\hsize=1.5\hsize}X
		}
		\toprule
		\textbf{DBMS-Workload} & \textbf{\#Knobs} &\textbf{Environment} \\ 
		\midrule
		\textbf{\envmysql} & 104  & \multirow{2}{=}{Virtual Machines - 16 vCPUs, 32GB RAM, 256GB SSD } \\
		\textbf{\envpg} & 70 & \\ 
		\midrule
		\textbf{\envsimsysbench} & 197  & \multirow{3}{=}{Open DBMS auto-tuning benchmark (Simulator) \cite{Bin2022}. } \\
		\textbf{\envsimjob} & 197 & \\ 
		\textbf{\envsimtpcc} & 100 & \\ 
		\bottomrule
	\end{tabularx}
\end{table}

\begin{table}[t]
	\centering
	\caption{Accuracy of NN-based simulators.}
	\label{tab:nnsim}
	\begin{tabular}{lcccccc}
		\toprule
		\multirow{2}{*}{\textbf{Method}} & \multicolumn{5}{c}{\textbf{$\text{R}^2$}} \\
		\multirow{2}{*}{} & SAINT & FT & AutoInt & ResNet & MLP \\ \midrule
		SYSBENCH & 0.803 & 0.7574 & 0.7252 & 0.3752 & 0.2862 \\ 
		JOB & 0.8164 & 0.8053 & 0.7174 & 0.3872 & 0.3186 \\ 
		TPCC & 0.9959 & 0.9995 & 0.9128 & 0.2281 & -0.2249 \\ 
		\bottomrule
	\end{tabular}
\end{table}

\textbf{Baseline optimizers.} We set up 21 baseline optimizers for DBMS auto-tuning that span vanilla and advanced GP-BOs, tree-ensemble BOs, BO with kernel regression, DNN and Parzen Density estimator surrogate models, reinforcement learning, and evolutionary optimization. (i) \textbf{\textit{Vanilla and advanced GP-BO baselines}.} We include the vanilla GP-BO, i.e., 
\textbf{(1) VBO}, as a basic baseline.
We also include advanced GP-BO variants, \textbf{(2) MixedBO}\cite{hutter2011, Bin2022}, \textbf{(3) HEBO}\cite{HEBO}, which wins the first place in NeurIPS 2020 black-box optimization challenge, \textbf{(4) HESBO}\cite{wang2016bayesian}, \textbf{(5) Turbo} \cite{eriksson2019scalable}, GP-BOs with Dot-Product, Absolute-Experiential and Mattern kernel, i.e., \textbf{(6) DP-BO}, \textbf{(7) AE-BO} and \textbf{(8) Mattern-BO} as baselines. Finally, we include a tree-structured BO scheme, \textbf{(9) LAMCTS} \cite{wang2020learning}, which uses MCTS (Monte Carlo Tree Search) to partition search space and fits multiple local GP-BO models for local configuration search.

(ii) \textbf{\textit{Tree-ensemble BO baselines}.} We include \textbf{(10) SMAC} \cite{lindauer2022smac3,li2021openbox, curino2020mlos,thornton2013auto} as a strong tree-ensemble BO baseline with an RF surrogate model. We also include other tree-ensemble BOs with GBDT surrogate models. \textbf{(11) NGB-BO} (Natural Gradient Boosting \cite{duan2020ngboost}) uses a multi-parameter boosting algorithm and the natural gradient technique to produce mean and interval estimates for GBDTs.\textbf{(12) PGBM-BO} (Probabilistic Gradient Boosting Machine) \cite{PGBM2021} approximates the leaf weights in a regression tree as a random variable and produces mean and interval estimates for GBDTs via stochastic tree ensemble update equations. \textbf{(13) VEGB} constructs mean and interval estimates for GBDT via a virtual ensemble technique\cite{malinin2021uncertainty}. The virtual ensemble uses truncated sub-models of a single multi-parameter GBDT model as elements of an ensemble and estimates both data and epistemic uncertainty. \textbf{(14) GBQRT-BO} \cite{skopt} adopts the classic quantile regression technique (with pinball loss) to produce mean and interval estimates for GBDTs.
\textbf{(15) SGBE-BO} (Stochastic Gradient Boosting Ensembles) \cite{malinin2021uncertainty}  is an ensemble of independent models generated via Stochastic Gradient Boosting (SGB). While each SGB weak learner is generated via LightGBM \cite{ke2017lightgbm}.

(iii) \textbf{\textit{Generalized BO baselines: BO with general statistical \& deep learning surrogate models}.} Besides GP and tree-ensembles, we also include baselines with more general ML-based surrogate models. This includes \textbf{(16) HORD}, BO with kernel regression and adopted by Alibaba's KeenTune auto-tuner \cite{keentune}, \textbf{(17) TPE}, BO with Parzen density estimator, 	\textbf{(18) DENN-BO}, BO with deep ensemble neural network, Monte Carlo dropout and probabilistic backpropagation and with comparable uncertainty quantification quality compared with BNN (Bayesian Neural Network)\cite{lakshminarayanan2017simple}. (iv) \textbf{\textit{RL baselines}} We include \textbf{(19) DDPG} (i.e., Deep Deterministic Policy Gradient algorithm), an RL (Reinforcement Learning)-based optimizer applied by CDBTune \cite{zhang2019end} and Qtune \cite{li2019qtune}. (v) \textbf{\textit{GA baselines.}} We include \textbf{(20) GA} \cite{GA2005} (Genetic Algorithm), a prevailing evolutionary computing  and optimizing algorithm, as baselines.

(iv) \textbf{OOB conformalized random forests \cite{johansson2014regression}}. See \cref{sec: compareconf}.

\textbf{Implementation}. 
For SMAC and MixedBO, we use the SMACV3's implementation \cite{lindauer2022smac3}.
For HEBO and DeepEnsemble, we use the HEBO's implementation \cite{HEBO}.
For GA and TPE, we use OpenBox’s implementation \cite{li2021openbox}.
For GBQRT-BO, we use the Scikit-Optimize's implementation \cite{skopt}.
For DDPG, we implement the neural network architecture used in CDBTune \cite{zhang2019end} 
with PyTorch \cite{NEURIPS2019_9015}.
For LlamaTune, TuRBO, LA-MCTS and HORD, we use their released implementations.
Lastly, we use the authors' implementation of NGBoost, PGBM, and Virtual Ensemble together 
with lightgbm \cite{ke2017lightgbm} and SMACV3 to implement 
NGB-BO, PGBM-BO, VEGB-BO, SGBE-BO, and \project.

\subsection{Evaluation on MySQL and PostgreSQL VMs}
\label{sec:eval_two_dbms}

\begin{figure}[t]
	\captionsetup{belowskip=0cm}
	\centering
	\begin{subfigure}[b]{0.47\textwidth}
		\centering
		\includegraphics[width=\textwidth]{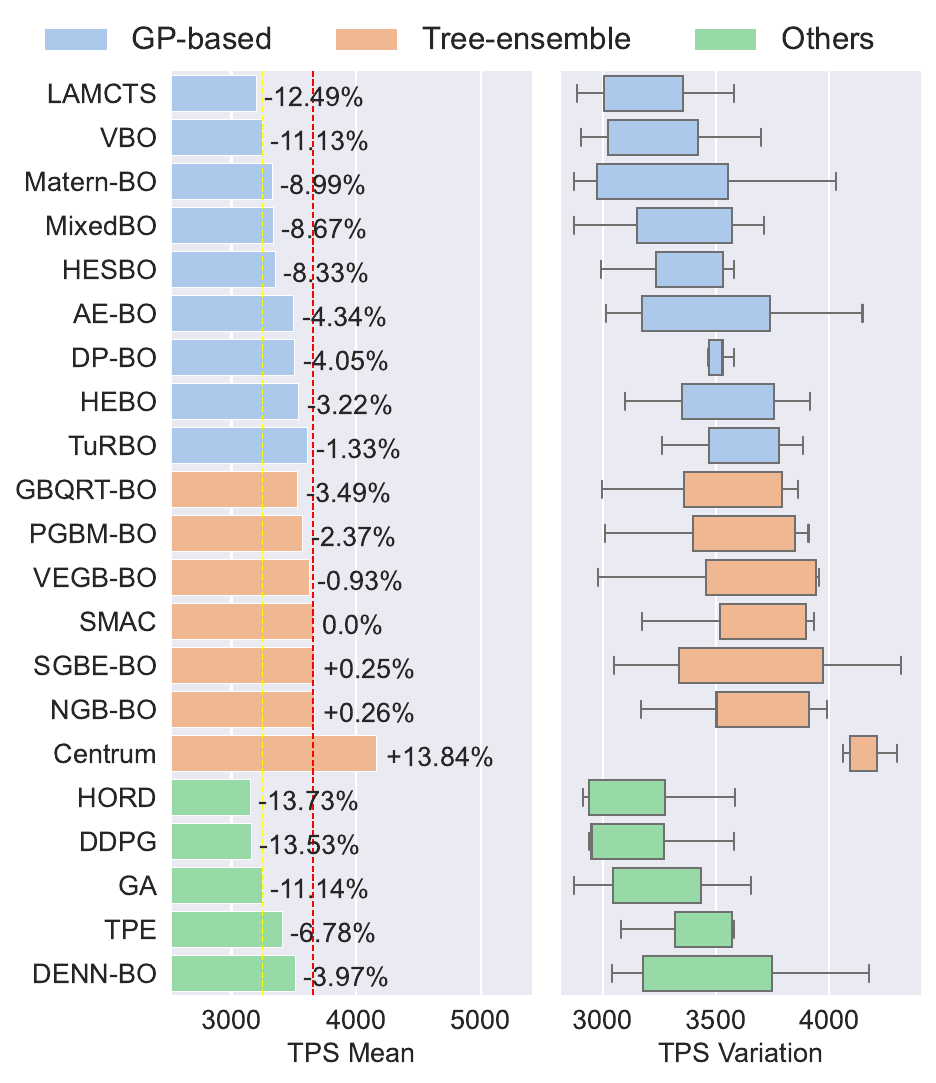}
		\caption{\envmysql}
		\label{fig:optim_mysql}
	\end{subfigure}
	\hfill
	\begin{subfigure}[b]{0.47\textwidth}
		\centering
		\includegraphics[width=\textwidth]{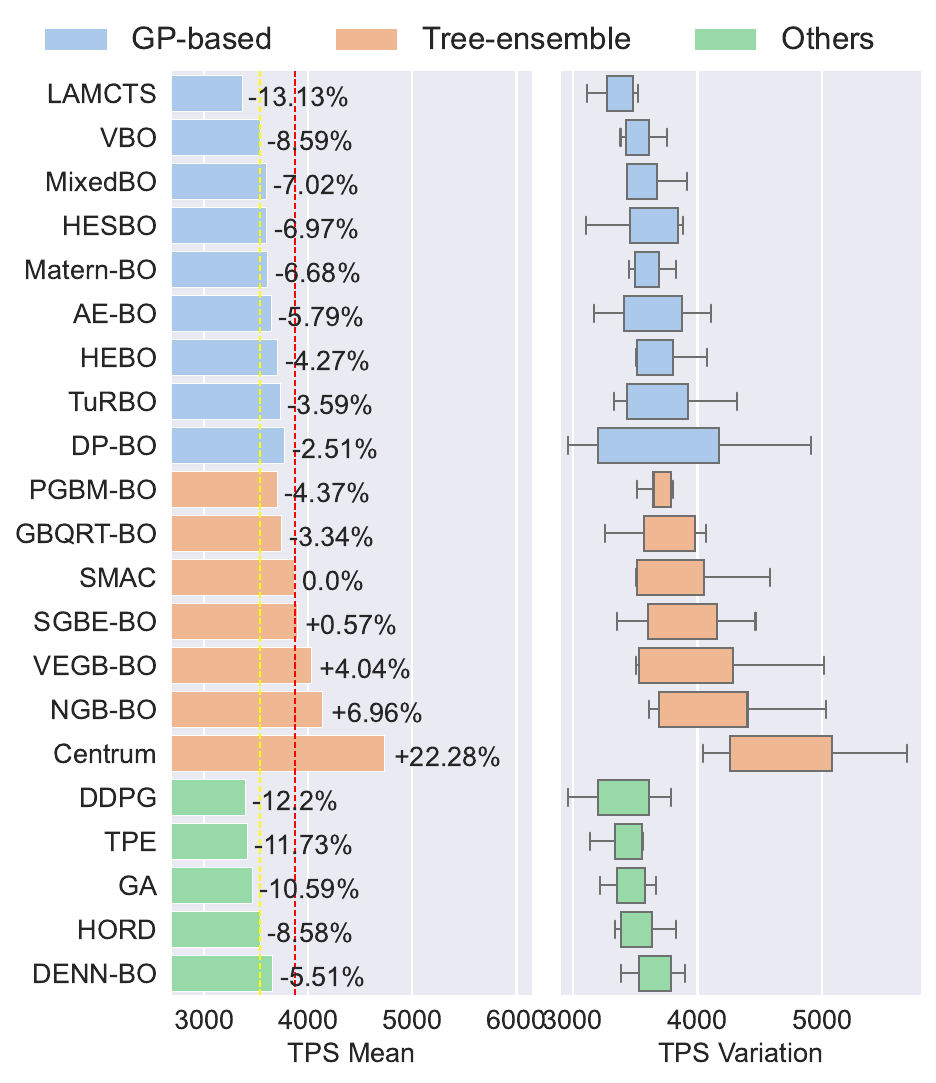}
		\caption{\envpg}
		\label{fig:optim_pg}
	\end{subfigure}
	\caption{Mean and variation of final tuned performance of MySQL-v8.0 and PostgreSQL-v10.5 over Sysbench.  Percentage numbers show relative improvements over SMAC. See entire tuning trajectories in \Cref{fig:eval_all_trajectory}.
	}
	\label{fig:perf_vm}
	\Description{The image shows a bar plot comparing TPS means and a box plot comparing TPS variations of different optimization methods on the MySQL8-SYSBENCH and PG10-SYSBENCH benchmarks. Optimizers are categorized into three groups: GP-based, Tree-ensemble, and Others.  Notable increases are seen in our proposed method (Centrum) of which the TPS mean exceeds that of SMAC by 13.84\% for MySQL-v8.0 and by 22.28\% for PostgreSQL-v10.5.}
\end{figure}

\textbf{\project outperforms GP-BOs, Tree-ensemble BOs, Generalized BOs, RL and GA.} \Cref{fig:perf_vm} shows \project achieves highest tuned performance compared with DBMS auto-tuners based on state-of-the-art optimizers. On average, \project produces 21.0\% and 28.9\% higher tuned throughput compared to auto-tuners based on 
advanced GP-BOs (21.6\% and 30.4\% higher),
tree-ensemble BOs (15.0\% and 21.5\% higher),
BOs with kernel regression (32.0\% and 33.8\% higher),
density estimator (22.1\% and 38.5\% higher) and
neural network surrogate models (18.5\% and 29.4\% higher),
reinforcement learning (31.6\% and 39.3\%) and
genetic algorithm (28.1\% and 36.8\% higher), 
in the \envmysql and \envpg experiments, respectively. 

\textbf{Tree-ensemble BOs systematically outperform GP-BOs, except for PGBM-BO and GBRT-BO.} 
\Cref{fig:perf_vm} shows tree-ensemble BOs, including \project (excluding \project), 
on average produce 8.6\% and 10.9\% (6.3\% and \%7.6) higher tuned throughput compared with GP-BO optimizers for \envmysql and \envpg, respectively. \project, NGB-BO, SGBE-BO, SMAC, and VEGB-BO individually outperform all GP-BO optimizers in the two experiments. However, not all tree-ensemble BOs are competitive, as GBQRT-BO is outperformed by DP-BO by TuRBO and HEBO in the two experiments, respectively; PGBM is outperformed by DP-BO, TuRBO, and HEBO, and by TuRBO, in the two experiments, respectively. GBQRT-BO uses vanilla gradient boosting regression tree (SKOPT\cite{skopt}'s implementation), which lacks advanced techniques such as over-fitting regularization and symmetric (balanced) trees and can cause inferior surrogate modeling accuracy (see \Cref{sec:eval_surrogate_accuracy}). PGBM-BO is limited to distributions using only location and scale to model the output, which in assumed Gaussian. Moreover, PGBM lacks modeling epistemic uncertainty. Both can undermine PGBM-BO's uncertainty quantification and its tuning efficacy  (see \Cref{sec:eval_surrogate_accuracy}).

\textbf{Except for \project, existing GBDT-based BOs do not consistently outperform SMAC.
	NGB-BO are at least on par with but can outperform SMAC in certain cases.}
\Cref{fig:perf_vm} shows \project significantly outperforms SMAC by 13.8\% and 22.3\% in optimizing throughput, in the \envmysql and \envpg experiments, respectively. However, no GBDT-based BOs further outperform SMAC in the \envmysql experiment. 
NGB-BO is on par with SMAC for three independent executions of the \envmysql setting but improves SMAC by 6.96\% in the \envpg experiment. 
SGBE-BO are on par with SMAC for both the experiments and GBQRT-BO and PGBM-BO are in general outperformed by SMAC. VEGB does not exhibit a stable improvement over SMAC.

\textbf{Non-stationary GP-BO's, particularly TuRBO, improves VBO more than other advanced GP-BOs.} \Cref{fig:perf_vm} shows 
TuRBO, DP-BO, and HEBO are strong GP-BO optimizers, which on average produce 8.4\% and 4.8\%, 5.4\% and 6.0\%, 6.3\% and 4.1\% higher throughput compared with the other optimizers in the GP-BO family,
for the \envmysql and \envpg experiments, respectively. 
TuRBO adopts multiple trust-region-supported local surrogate models to piecewisely fit the DBMS performance model, which can be a non-smooth, non-continuous, complex surface (cf.,\Cref{fig:motivation_non_smooth0}). DP-BO adopts a non-stationary Dot-Product kernel that adopts varying interdependence and varying covariance between configuration knobs (cf.,\Cref{fig:motivation_lengthscales0}). TuRBO and DP-BO are both non-stationary GP-BOs. HEBO stabilizes non-stationary variance and reifies non-Gaussianity, by using Box-Cox and Yeo-Jonhson transformations to transform DBMS-performance measurements. Results of \envmysql and \envpg suggest lifting and resolving the stationarity assumption help most to improve beyond VBO, compared to resolving other restrictive assumptions. Other advanced GP-BOs, MixedBO, and HESBO slightly improve VBO while LAMCTS fails. LAMCTS trains a large number of classifiers (exponential to the depth of the Monte Carlo search tree) to partition space, where we find the classifiers can be under-fitted provided a small number of DMBS performance samples.

\textbf{Selecting kernel with reduced smoothness improves VBO.} By switching the kernel of VBO from RBF to AE (Absolute Exponential) and Matern ($\gamma$=2.5), \Cref{fig:perf_vm} shows
AE-BO and Mattern-BO improve VBO by 7.6\% and 2.4\%, 3.0\% and 2.1\%, for the two experiments, respectively.
RBF kernels are accused of being overly smooth and unrealistic for modeling many physical processes. Results coincide with critics as RBF is smoother than Matern ($\gamma$=2.5), which is smoother than AE, and shows an increasing tuned performance from RBF to Matern to AE. Though kernel tuning helps improve VBO, it is not on par with advanced GP-BO and tree-ensemble BO.

\textbf{Generalized BOs, RL and GA are systematically outperformed by GP-BOs and Tree-ensemble BOs.} \Cref{fig:perf_vm} shows 
HORD, DDPG, GA, TPE, and DENN-BO yield on average 3.1\% and 3.4\% and 10.8\% and 13.0\% lower throughput to the GP-BOs and the tree-ensemble BOs, 
for the two experiments, respectively.

\subsection{Evaluation on Open Benchmark Simulators}
\label{sec:eval_three_sim}

\begin{figure}[t!]
	\centering
	\begin{subfigure}[b]{0.48\textwidth}
		\centering
		\includegraphics[width=\textwidth]{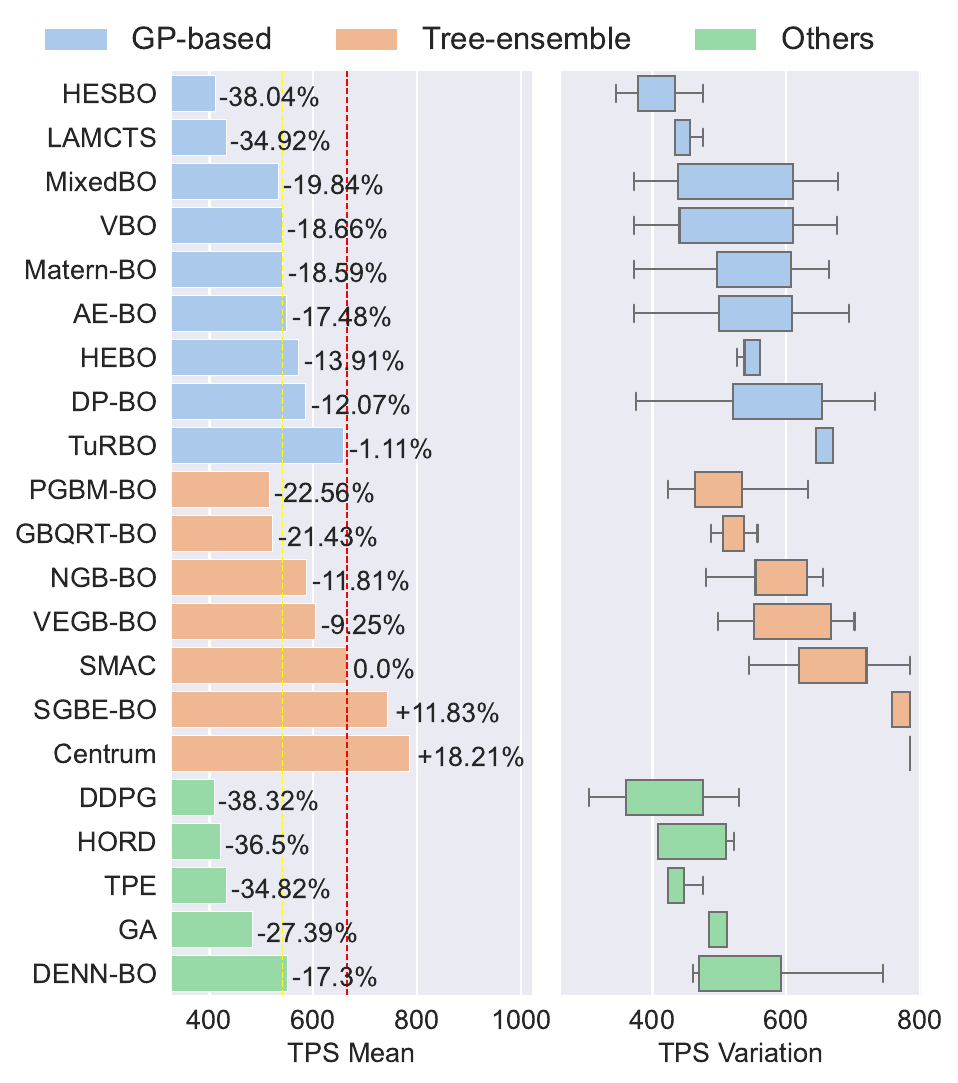}
		\caption{\envsimsysbench}
		\label{fig:optim_saint_sysbench}
	\end{subfigure}
	\hfill
	\begin{subfigure}[b]{0.47\textwidth}
		\centering
		\includegraphics[width=\textwidth]{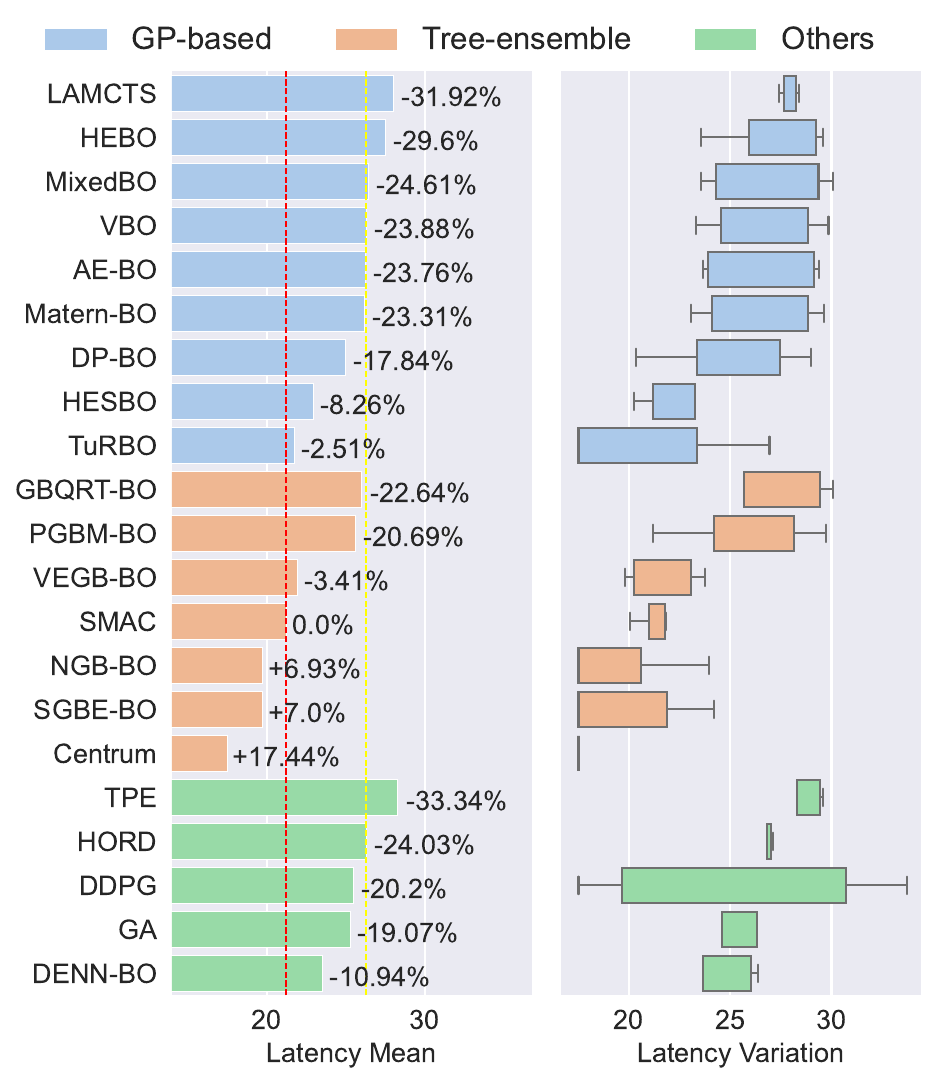}
		\caption{\envsimjob}
		\label{fig:optim_saint_job}
	\end{subfigure}
	\hfill
	\begin{subfigure}[b]{0.48\textwidth}
		\centering
		\includegraphics[width=\textwidth]{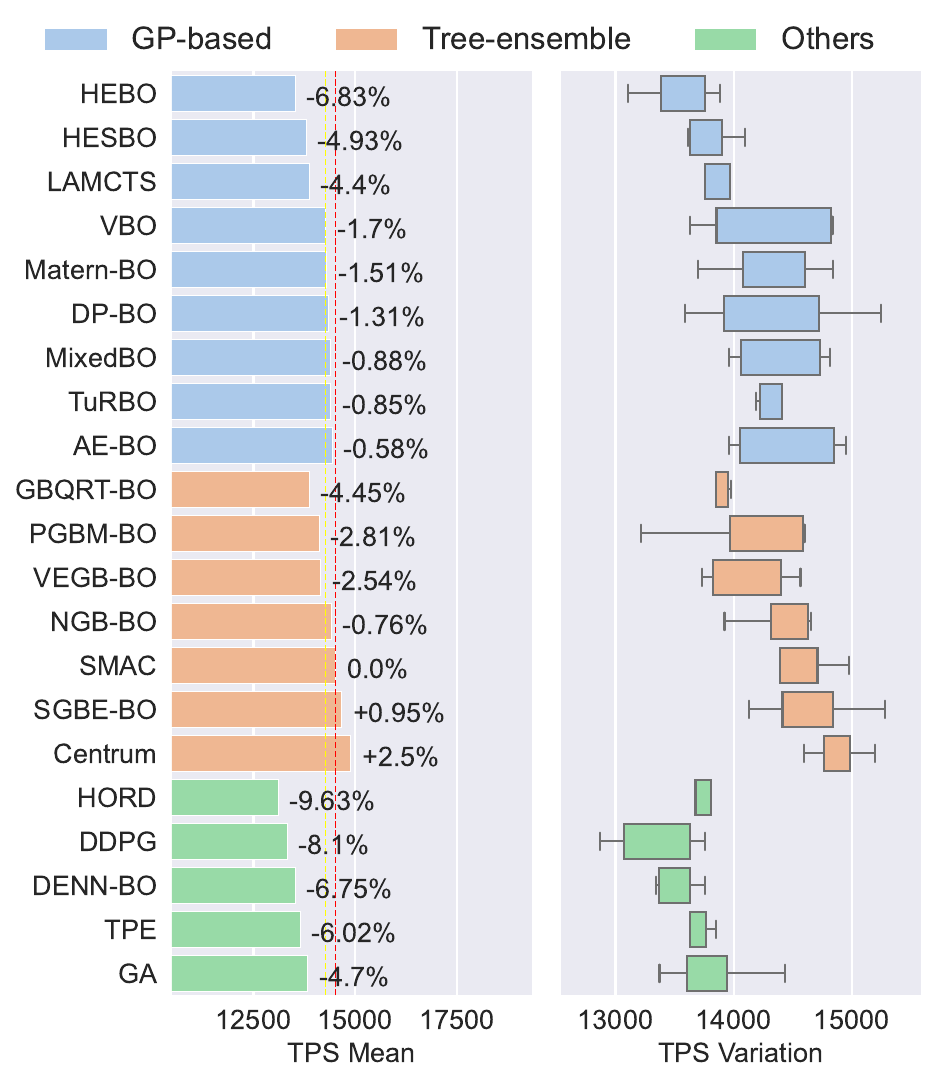}
		\caption{\envsimtpcc}
		\label{fig:optim_saint_tpcc}
	\end{subfigure}
	\caption{Mean and variation of final tuned performance of MySQL-v5.7 over Sysbench, Job and TPCC.  Percentage numbers show relative improvements over SMAC. See entire tuning trajectories in \Cref{fig:eval_all_trajectory}.}
	\label{fig:perf_saint}
	\Description{The image shows a bar plot comparing performance (TPS or Latency) means and a box plot comparing TPS variations of different optimization methods on the MySQL5-SYSBENCH, MySQL5-JOB, and MySQL5-TPCC benchmarks. Optimizers are categorized into three groups: GP-based, Tree-ensemble, and Others.  Notable increases are seen in our proposed method (Centrum) of which the performance mean exceeds that of SMAC by 18.2\% for MySQL5-SYSBENCH, by 17.44\% for MySQL5-JOB, and by 2.5\% for MySQL5-TPCC.}
\end{figure}

\Cref{fig:perf_saint} show the results of three simulation experiments, \envsimsysbench, \envsimjob, and \envsimtpcc, on open benchmark simulators \cite{Bin2022} (see \Cref{sec:experiment_method} for our modification). The simulation experiments primarily differ from the physical VM experiments in two aspects. First, the simulations have a 1.97$\times$-2.81$\times$ larger configuration parameter space (197 knobs compared to 100 and 70 knobs). Learning and optimizing within a high-dimensional space is more challenging. Second, the number of workloads extends from one to three, Sysbench, JOB, and TPCC, which facilitates reducing observation and conclusion biases in \Cref{sec:eval_two_dbms}.

Results of optimizing simulator throughput in \envsimsysbench and simulator latency in \envsimjob, are shown in \Cref{fig:optim_saint_sysbench,fig:optim_saint_job}, respectively,
which mostly coincides with that of the physical experiments in \Cref{sec:eval_two_dbms}. 

First, 
results reassert \project on average outperforms
GP-BOs by 22.3\% (31.6\%),
other tree-ensemble BOs by 15.0\% (21.7\%),
generalized BOs by 24.0\% (32.7\%),
RL by 31.6\% (31.3\%) and
GA by 28.1\% (30.6\%),
in producing higher (lower) tuned simulator throughput (latency).
Independent experiments show \project's consistent improvement in DBMS auto-tuning over existing baseline optimizers.

Second, results reconfirm prior findings for tree-ensemble BOs.  Tree-ensemble BOs generally outperform GP-BOs, except for PGBM-BO and GBQRT-BO. 
SGBE-BO significantly outperforms the other baselines including SMAC,
but still with an 11.9\% (12.6\%) gap in tuned throughput (latency) to match \project;
NGB-BO is on par with SMAC with comparable tuned throughput, but yields a 6.93\% lower tuned latency; 
VEGB and SMAC are on par for both simulation experiments; PGBM-BO and GBQRT-BO remain the least performant tree-ensemble optimizers and are outperformed by GP-BOs.

Third, TuRBO, and DP-BO continue to significantly improve GP-BOs, which reasserts that adopting non-stationary GP-BOs improves beyond VBO most. However, other advanced BO techniques, including AE-BO, Matter-BO, HEBO, HESBO, MixedBO, and LAMCTS do not consistently improve VBO across the two experiments of \envsimsysbench and \envsimjob.

\begin{figure}[t!]
	\centering
	\includegraphics[width=0.5\textwidth]{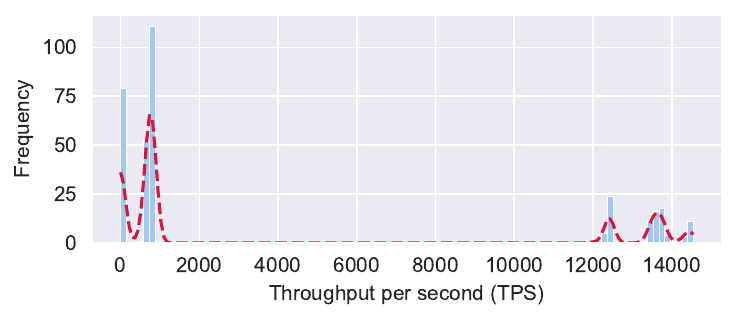}
	\caption{Diagnosing irregularities in DBMS-throughput distribution for benchmark datasets (MySQL5-TPCC).}
	\label{fig:perf_muti_modal}
	\Description{A histogram plot of the TPS performance data for generating the MYSQL5-TPCC benchmark. A majority of samples lie in the very poor region (TPS<2000) while the other samples concentrate on a remarkably higher performed region (TPS>11000). Such extreme multi-modal distribution causes all optimizers reach quasi-optimal solution easily.}
\end{figure}

\Cref{fig:optim_saint_tpcc} shows the evaluation results for experiment \envsimtpcc, which deliver striking different results compared to the four previous experiments \envmysql, \envpg, \envsimsysbench, and \envsimjob. \project delivers the best-tuned simulator throughput, but only 2.5\% higher than SMAC and 4.2\% better than VBO. 
Specifically, all GP-BO and tree-ensemble optimizers yield comparable tuned throughput, within a maximal 5\% gap. However, the ranking of varied optimizers resembles that in the previous four experiments. \project and SGBE-BO (and PGBM and GBQRT-BO) remain the best performant (least performant) tree-ensemble BOs and TurBo and DP-BO (LAMCTS) are performant (non-performant) GP-BOs. Finally, generalized BOs, RL, and GA remain unable to match GP-BOs and tree-ensemble BOs. To further parse such degenerate behavior of optimizers, we find that the collected DBMS throughput samples in related datasets (of the open benchmark) follow a simple tri-modal distribution. Most of the samples are ``unsafe'' configurations with collapsed throughput near zero. The other samples are all quasi-optimal configurations. There are no intermediate configurations in between and complex performance prediction reduces to a binary classification problem. Thus, all optimizers can reach a quasi-optimal solution easily.

\begin{figure*}[t!]
	\centering
	\includegraphics[width=0.98\textwidth]{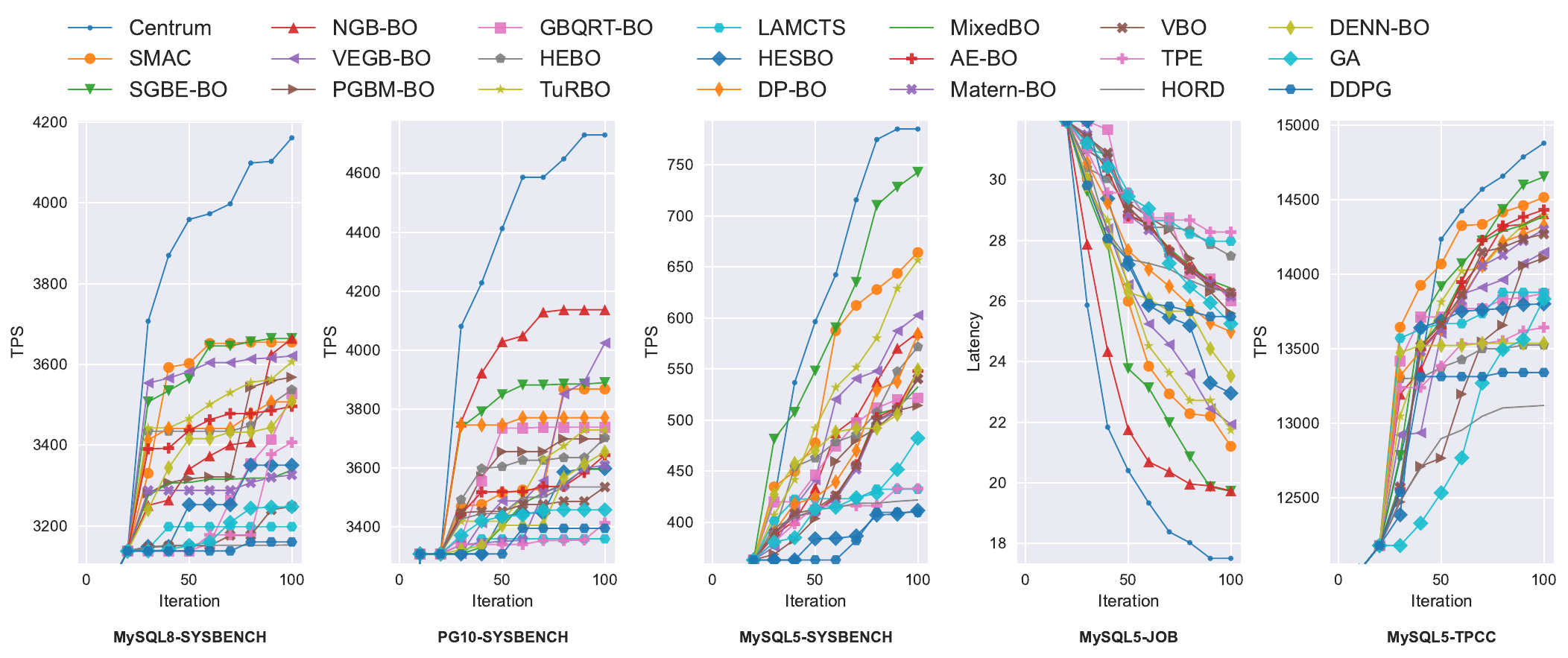}
	\caption{Tuning trajectory for physical (averaged across 3 runs) and simulation (averaged across 5 runs) experiments.}
	\label{fig:eval_all_trajectory}
	\Description{There are five line graphs of which each compare the performance of 21 different optimization methods over 100 iterations under the same benchmark. The graph indicates that our proposed method (Centrum) achieves the best final-tuned performance as well as the fastest improvement rate.}
\end{figure*}

\setlength{\tabcolsep}{2.5pt} 
\begin{table*}[tb]
	\centering
	\caption{Time-to-optimum in the number of iterations. \tto{10} and \tto{20}  denotes the number of iterations to reach within 10\% and 20\% of the targeted optimum respectively. ``\textbackslash'' denotes not reaching the optimum. \tsen denotes the extrapolated num. iterations to reach the optimum using estimated Theil-Sen's slope.}
	\label{tab:efficiency_stats}
	\scriptsize
	\begin{tabular}{llllllllllllllllc}
		\toprule
		\multirow{2}{*}{\textbf{Method}} & 
		\multicolumn{3}{c}{\textbf{\envmysql}} & \multicolumn{3}{c}{\textbf{\envpg}} & 
		\multicolumn{3}{c}{\textbf{\envsimsysbench}} & \multicolumn{3}{c}{\textbf{\envsimjob}} & 
		\multicolumn{3}{c}{\textbf{\envsimtpcc}} & 
		\multirow{2}{*}{\shortstack{\textbf{Average} \\ \textbf{\tsen}}}\\ 
		\cline{2-4}\cline{5-7}\cline{8-10}\cline{11-13}\cline{14-16}
		\multirow{2}{*}{} & 	
		\textbf{\tto{10}} & \textbf{\tto{20}}  & \textbf{\tsen} & 
		\textbf{\tto{10}} & \textbf{\tto{20}}  & \textbf{\tsen} & 
		\textbf{\tto{10}} & \textbf{\tto{20}}  & \textbf{\tsen} & 
		\textbf{\tto{10}} & \textbf{\tto{20}}  & \textbf{\tsen} & 
		\textbf{\tto{10}} & \textbf{\tto{20}}  & \textbf{\tsen} & 
		\\
		\midrule
		Centrum & 27 & 21 & 82 & 27 & 27 & 134 & 49 & 37 & 54 & 39 & 31 & 186 & 55 & 42 & 283 & 148 \\ 
		NGB-BO & 94 & 81 & 108 & 34 & 29 & 260 & \textbackslash & 78 & 105 & 46 & 40 & 191 & 80 & 72 & 432 & 219 \\ 
		SMAC & 58 & 39 & 100 & \textbackslash & 71 & 298 & 59 & 54 & 78 & 91 & 56 & 207 & 59 & 51 & 475 & 232 \\ 
		SGBE-BO & 56 & 33 & 132 & \textbackslash & 33 & 582 & 58 & 41 & 66 & 70 & 46 & 187 & 73 & 62 & 312 & 256 \\ 
		TuRBO & \textbackslash & 61 & 212 & \textbackslash & \textbackslash & 406 & 81 & 57 & 85 & 98 & 57 & 217 & 90 & 72 & 369 & 258 \\ 
		VEGB-BO & \textbackslash & 29 & 313 & 89 & 78 & 277 & 89 & 59 & 95 & 94 & 59 & 239 & \textbackslash & 93 & 403 & 265 \\ 
		PGBM-BO & \textbackslash & 78 & 120 & \textbackslash & \textbackslash & 445 & \textbackslash & \textbackslash & 133 & \textbackslash & \textbackslash & 403 & \textbackslash & 93 & 401 & 300 \\ 
		GBQRT-BO & \textbackslash & 95 & 151 & \textbackslash & 69 & 385 & \textbackslash & 83 & 137 & \textbackslash & \textbackslash & 375 & \textbackslash & \textbackslash & 741 & 358 \\ 
		HESBO & \textbackslash & \textbackslash & 198 & \textbackslash & \textbackslash & 543 & \textbackslash & \textbackslash & 358 & \textbackslash & 77 & 253 & \textbackslash & \textbackslash & 479 & 366 \\ 
		AE-BO & \textbackslash & \textbackslash & 276 & \textbackslash & \textbackslash & 725 & \textbackslash & 93 & 133 & \textbackslash & \textbackslash & 397 & 79 & 63 & 351 & 376 \\ 
		MixedBO & \textbackslash & \textbackslash & 648 & \textbackslash & \textbackslash & 543 & \textbackslash & 93 & 137 & \textbackslash & \textbackslash & 402 & 88 & 62 & 367 & 419 \\ 
		Matern-BO & \textbackslash & \textbackslash & 783 & \textbackslash & \textbackslash & 552 & \textbackslash & 92 & 137 & \textbackslash & \textbackslash & 389 & \textbackslash & 76 & 387 & 450 \\ 
		GA & \textbackslash & \textbackslash & 356 & \textbackslash & \textbackslash & 1311 & \textbackslash & \textbackslash & 200 & \textbackslash & 99 & 361 & \textbackslash & \textbackslash & 432 & 532 \\ 
		VBO & \textbackslash & \textbackslash & 736 & \textbackslash & \textbackslash & 1338 & \textbackslash & 94 & 137 & \textbackslash & \textbackslash & 388 & \textbackslash & 64 & 382 & 596 \\ 
		HEBO & \textbackslash & 99 & 396 & \textbackslash & \textbackslash & 630 & \textbackslash & 86 & 126 & \textbackslash & \textbackslash & 635 & \textbackslash & \textbackslash & 1511 & 660 \\ 
		DENN-BO & \textbackslash & \textbackslash & 160 & \textbackslash & \textbackslash & 513 & \textbackslash & 91 & 155 & \textbackslash & 82 & 293 & \textbackslash & \textbackslash & 2796 & 783 \\ 
		DP-BO & \textbackslash & \textbackslash & 241 & \textbackslash & 23 & 3315 & \textbackslash & 78 & 115 & \textbackslash & 90 & 340 & 97 & 73 & 414 & 885 \\ 
		DDPG & \textbackslash & \textbackslash & 2148 & \textbackslash & \textbackslash & 1665 & \textbackslash & \textbackslash & 332 & \textbackslash & \textbackslash & 378 & \textbackslash & \textbackslash & 701 & 1045 \\ 
		TPE & \textbackslash & \textbackslash & 499 & \textbackslash & \textbackslash & 3131 & \textbackslash & \textbackslash & 352 & \textbackslash & \textbackslash & 625 & \textbackslash & \textbackslash & 1036 & 1129 \\ 
		LAMCTS & \textbackslash & \textbackslash & 822 & \textbackslash & \textbackslash & 3315 & \textbackslash & \textbackslash & 563 & \textbackslash & \textbackslash & 538 & \textbackslash & \textbackslash & 1368 & 1321 \\ 
		HORD & \textbackslash & \textbackslash & 7159 & \textbackslash & \textbackslash & 807 & \textbackslash & \textbackslash & 581 & \textbackslash & \textbackslash & 405 & \textbackslash & \textbackslash & 766 & 1944 \\
		\bottomrule
	\end{tabular}
\end{table*}

\begin{figure}[htb]
	\centering
	\begin{subfigure}[b]{0.9\textwidth}
		\centering
		\includegraphics[width=\linewidth]{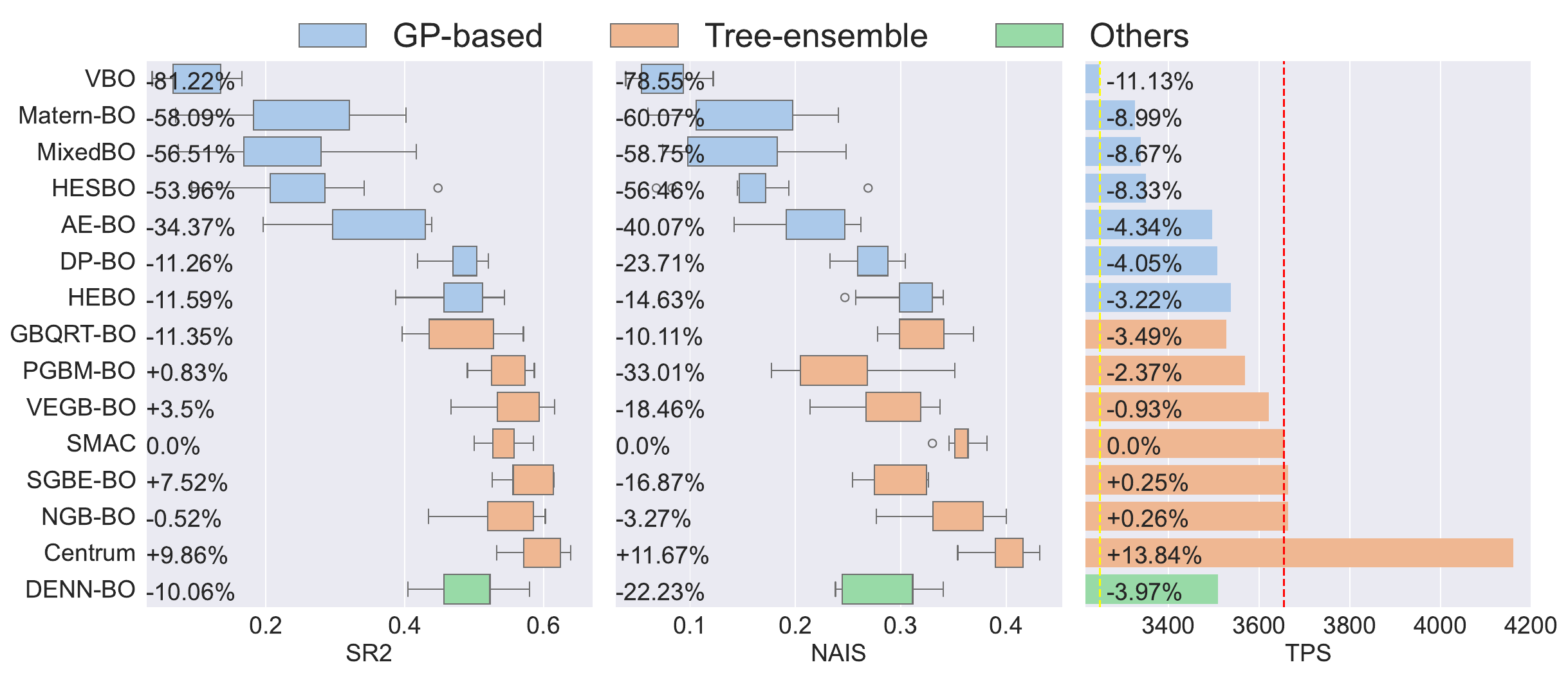}
		\caption{\envmysql}
		\label{fig:perf_zoomin_mysql}
		\Description{The surrogate evaluation of Centrum and compared optimizers on tuning MySQL under SYSBENCH workload.}
	\end{subfigure}
	\hfill
	\centering
	\begin{subfigure}[b]{0.9\textwidth}
		\centering
		\includegraphics[width=\linewidth]{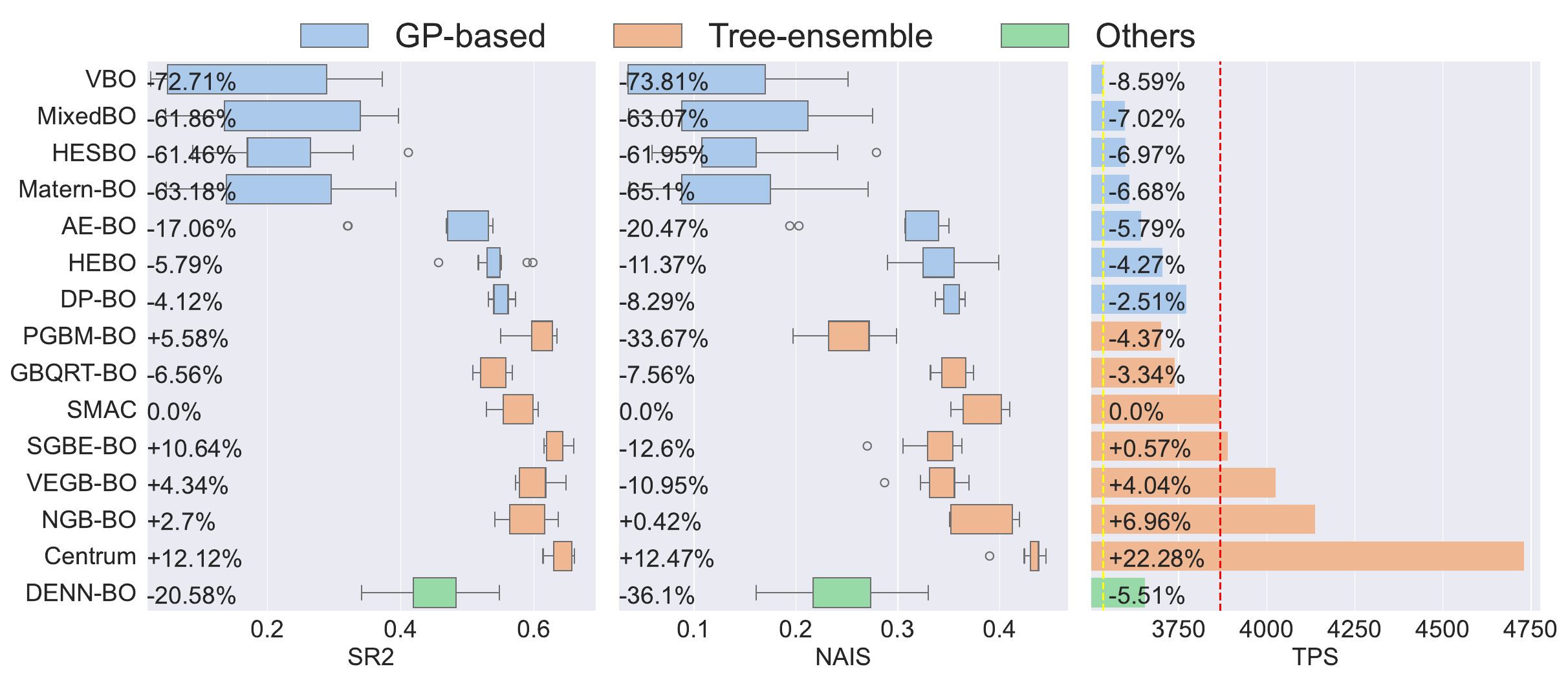}
		\caption{\envpg}
		\label{fig:perf_zoomin_pg}
		\Description{The surrogate evaluation of Centrum and compared optimizers on tuning PostgreSQL under SYSBENCH workload.}
	\end{subfigure}
	\label{fig:perf_zoomin}
	\caption{Point-prediction accuracy ($\text{R}^2$) and interval-prediction accuracy ($\text{NAIS}$) estimated from the \envmysql and \envpg experiments. Percentage numbers show relative improvements over SMAC.}
	\Description{Throughput performance evaluations of Centrum and compared optimizers on MySQL and PostgreSQL along with their underlying surrogates' point-prediction accuracy (SR2) and the interval-prediction (NAIS) respectively. These figures show that Centrum achieves better both point and interval accuracy, and thereby better final-tuned TPS performance.}
\end{figure}

\subsection{Evaluation of Tuning Efficiency}
\label{sec:efficiency}

\textbf{Theil-Sen slope estimator to measure DBMS auto-tuning efficiency.} Prior work on DBMS auto-tuning measures the tuning efficiency of an optimizer by \tto{x} (time to optimal), the number of iterations required for a DBMS auto-tuner to reach a target performance $y_o$ \cite{kanellis2022llamatune,zhang2022towards}. Provided a fixed number of iterations, $y_o$ is usually set as the maximal performance seen across all optimizers under the budget. However, some optimizers are unable to reach $y_o$ over the iterations, this excludes them from efficiency evaluation. To address this problem, people use a small fraction $x$\% (e.g., 10\%) to lower the optimum to a quasi-optimum(1-$x$\%)$y_o$, which can potentially allow more optimizers to reach it for efficiency evaluation. \Cref{tab:efficiency_stats} shows \tto{10} and \tto{20} for the physical and simulation experiments in \Cref{sec:eval_two_dbms,sec:eval_three_sim}. However, there is still a large fraction of optimizers that can not reach the 80\% target. Rather than continue lowering to a loose target, we propose using \textbf{Theil-Sen’s slope estimator $\beta_{Sen}$}\cite{sen1968estimates} to robustly measure the average performance increasing rate for each optimizer. Then using the slope, we can extrapolate the number of steps required to reach the uncompromised optimum solution $y_o$ by $\text{T}_{\text{sen}}=y_o/\beta_{Sen}$ for all optimizers (see \Cref{tab:efficiency_stats}).

\noindent\textbf{\project is the fastest optimizer w.r.t \tto{10}, \tto{20}, and \tsen}. \Cref{tab:efficiency_stats}) shows \project ranks the first for all five experiments with average \tto{10}=39, \tto{20}=32 and \tsen=148. \tsen allows us to compare \project with all baseline optimizers;  shows that \project is 1.84X faster than the other tree-ensemble methods, 4.01X faster than GP-BOs,
and 7.35X faster than generalized BOs, RL, and GA in tuning DBMS to the target performance. \Cref{fig:eval_all_trajectory} visualizes the tuning trajectory for the physical and simulation experiments. It is striking that \project achieves superior tuning speed-ups compared with baseline optimizers.

\noindent\textbf{Tree-ensemble BOs are faster than GP-BOs w.r.t. \tsen. TuRBO is comparably fast to Tree-ensemble BOs}. \Cref{tab:efficiency_stats}) shows that existing tree-ensemble methods are 2.18X faster than GP-BOs, where NGB-BO, SMAC,  and SGBE-BO achieve the highest speedups. TuRBO, as a strong non-stationary GP-BO, shows a stunningly comparable speedup to the leading tree-ensemble scheme and achieves 2.46x higher speedup than remaining GP-BO methods.

\subsection{Evaluation of Surrogate Modeling Accuracy}\label{sec:eval_surrogate_accuracy}

\setlength{\tabcolsep}{4pt} 
\begin{table}[!ht]
\centering
\caption{Average probabilities of concordance estimated from \envmysql and \envpg. BOTH stands for the conjunction of (\sr, \nais)}  
\label{tab:poc}
\begin{tabular}{lcccccc}
	\toprule
    Methods & $P_{CC}$(TPS,\sr) & $P_{CC}$(TPS,\nais) & $P_{CC}$(TPS,BOTH) \\ \midrule
	All & 0.8856 & 0.8906 & 0.9430 \\ 
	Tree-ensemble & 0.6906 & 0.7860 & 0.9050 \\ 
	\bottomrule
\end{tabular}
\end{table}

We explain the performance differences between DBMS-tuning optimizers by parsing their quality of exploitation and exploration. 
We evaluate the quality of exploitation and exploration with the point prediction and the interval prediction accuracy of the surrogate model, which are measured by \sr and \nais, respectively, as introduced in \Cref{sec:finetune}. We compute \sr and \nais for individual optimizers and visualize them in \Cref{fig:perf_zoomin} for the physical experiments in \Cref{sec:eval_two_dbms}. \sr and \nais metrics for the simulation experiments and some optimizers are omitted to save space.

Specifically, we collect the dataset used to train individual optimizer's surrogate model, which is the trajectory of (configuration, TPS) pairs within their entire tuning life-cycle. However, each optimizer has its own inductive bias and targeted areas in the configuration space. The collected trajectory data usually lie in different areas and have different noise distributions. Such disparity can cause unfair comparisons between optimizers' surrogate models as there is unparalleled difficulty in fitting differently distributed data in different areas. To resolve such bias and guarantee fairness, we merge the collected trajectory data across all optimizers and form a unified dataset and observation of the unknown DBMS performance surface of DBMS. We then apply a 10/90 train-test split to train and test individual optimizers' surrogate models. We compute and report the averaged \sr and \nais over 10 evaluations on random train-test splits. 

\textbf{Evaluating the decisiveness of \sr and \nais over final tuned performance.} We first evaluate if \sr and \nais and decisive to the final tuned performance for individual optimizers. We need to measure the concordance between a statistical indicator and the final tuned throughput; that is, if optimizer $A$'s \sr, \nais, or both is higher than that of optimizer $B$, the likelihood that $A$ also has a higher tuned-TPS than $B$. A higher concordance indicates a more significant decisiveness of the indicators. We compute the probability of concordance measure \cite{denuit2006actuarial} $P_{CC}$(TPS,$M$) between final tuned-TPS and the indicator $M$, where $M$ can be \sr, \nais, or BOTH,  i.e., (\sr, \nais), as shown in \Cref{tab:poc}. First, results show on average for all optimizers, \sr or \nais is individually decisive to final tuned-TPS, while their joint decisiveness is high (approach upper-bound one). Second, for tree-ensemble optimizers' surrogate models, neither \sr nor \nais along is a decisive indicator of final-tuned TPS, but they jointly show a high decisiveness. Third, \nais on average has a higher decisiveness than \sr.

\textbf{\project outperforms other tree-ensemble BOs and GP-BOs w.r.t. \sr and \nais}.
\Cref{fig:perf_zoomin} shows \project has the highest \sr and \nais among all optimizers. In particular,
compared with tree-ensemble BOs, \project on average increases \sr and \nais by 9.5\% and 27.6\% respectively, linking to 18.27\% TPS improvement. When compared with GP-BOs, \project increases \sr and \nais by 92.6\% and 105.7\% respectively, linking to 26.21\% TPS improvement. Overall, the leading \sr and \nais translate to high DBMS auto-tuning effectiveness and efficiency for \project.

\textbf{Existing GBDT-based surrogate models benefit from high \sr but suffer from lower \nais.} \Cref{fig:perf_zoomin} shows GBDT-based surrogate models NGB, SGBE, VEGB, and PGBM have 4.87\% higher \sr compared to compared to SMAC's RF surrogate model.
GBQRT-BO is an exception that has on average 8.96\% lower \sr compared to SMAC. However, the \nais's of GBDT-based surrogate models are remarkably lower (-3.97\%) than that of RF. In particular, PGBM's \nais is significantly lower (33.34\% lower) than that of RF. Such analyses explain the inferiority of GBQRT-BO and PGBM-BO compared to other tree-ensemble BOs and validate that inaccurate uncertainty estimation is a limit factor for existing GBDT-based BO schemes.

\textbf{Tree-ensemble BOs and non-stationary GP-BOs.}
\Cref{fig:perf_zoomin} shows, on average, tree-ensemble BOs has 78.37\% higher \sr and 67.5\% higher \nais compared to GP-BOs, linking to a 9.48\% improvement of TPS over GP-BOs. GP-BOs with stationary kernels including VBO, MixedBO, HESBO, and Matern-BO all exhibit significantly deficient \sr and \nais, which can explain their inferior tuned throughput compared to non-stationary BOs such as DP-BO. 

Overall, the differences in final-tuned performance can be explained by the differences in \sr and \nais to a high precision. It implies improving \sr and \nais, especially \nais is the key to achieving effective and efficient DBMS auto-tuning.

\subsection{Comparison of Centrum against Out-of-Bag-Conformalized Methods}
\label{sec: compareconf}

\begin{figure}[htb]
	\centering
	\includegraphics[width=0.55\textwidth]{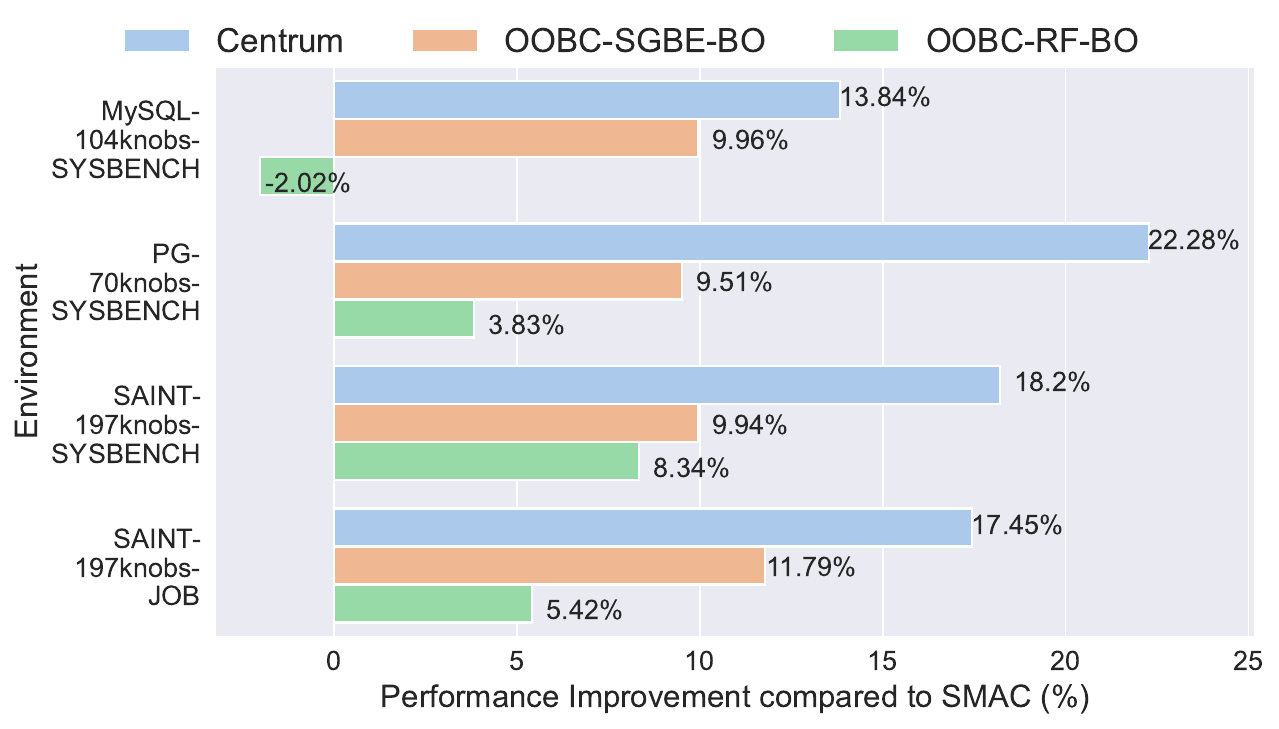}
	\caption{Relative improvements over SMAC for \project, OOBC-RF and OOBC-SGBE.}
	\label{fig:perf_oob}
	\Description{The bar chart compares the relative performance improvements over SMAC for Centrum, OOBC-RF and OOBC-SGBE across 4 benchmarks. It indicate that Centrum has significantly higher improvement compared to existing conformal methods or applications.}
\end{figure}

We further compare \project against out-of-bag-conformalized (OOBC)-based BO \cite{linusson2020efficient} including OOBC-RF-BO and OOBC-SGBE-BO.
For OOBC-RF, we implement the surrogate model as the RFoa model \cite{johansson2014regression} with scikit-learn \cite{scikit-learn} and SMACV3\cite{lindauer2022smac3}, which consists of a random-forest-based surrogate and an interval estimation produced by the out-of-bag conformal score and an ERC difficulty measurement. 
We apply the same procedure to SGBE to create OOBC-SGBE-BO.
\Cref{fig:perf_oob} shows the relative improvement over SMAC for \project, OOBC-RF-BO and OOBC-SGBE-BO.
We observe that OOBC-RF-BO increases the average DBMS performance by 3.89\% over SMAC,
and OOBC-SGBE-BO further lifts the average performance increment to 10.30\%.
This result highlights that both the OOB conformal method and the SGBE-based surrogate model
are effective for improving existing tree-ensemble-based methods.
Moreover, \project outperforms OOBC-RF-BO and OOBC-SGBE-BO by 14.05\% and 7.64\% on average, respectively, indicating the effectiveness of the proposed generalized locally adaptive conformal method against existing conformal methods.

\subsection{Ablation Study of \project}

\begin{figure}[htb]
	\centering
	\includegraphics[width=0.55\textwidth]{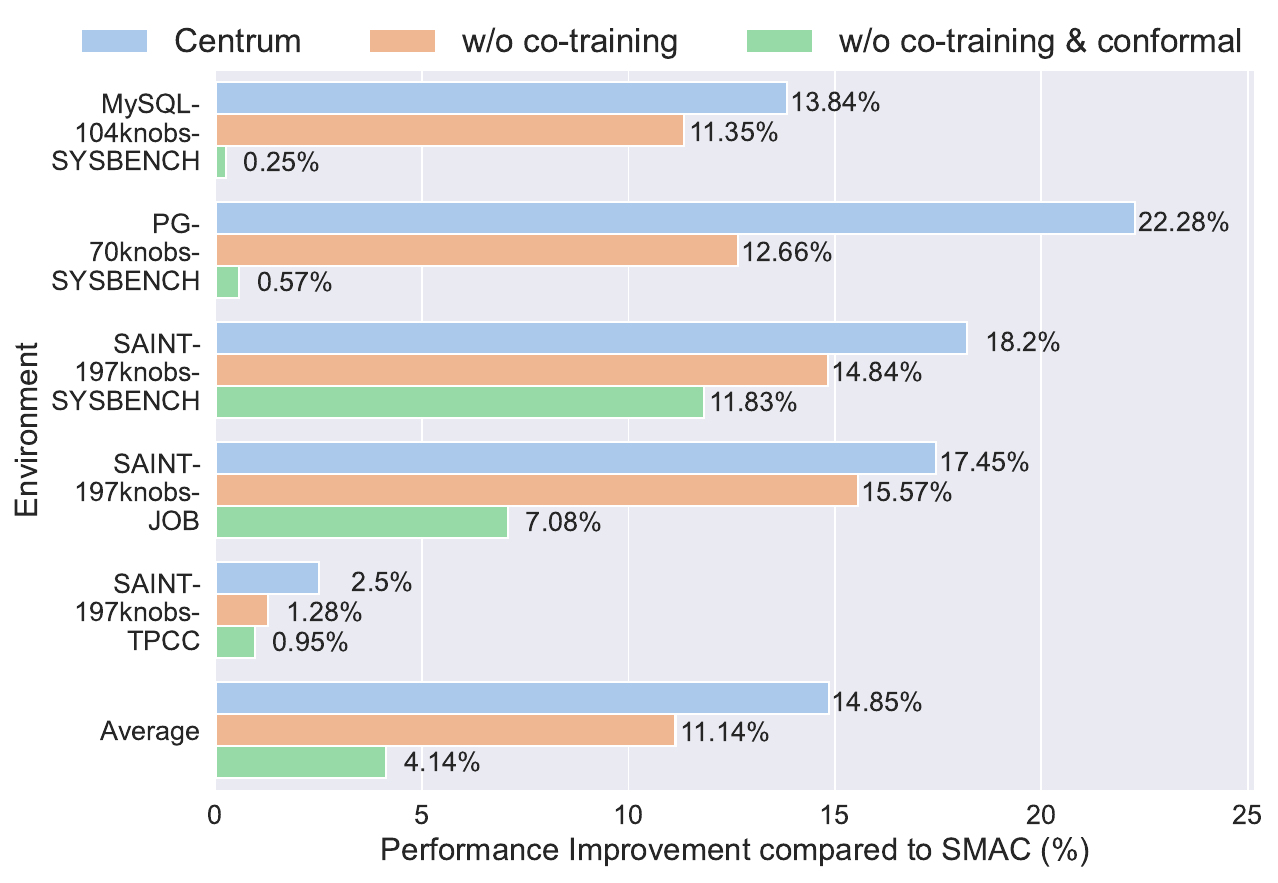}
	\caption{Ablation study with relative improvements over SMAC.
	}
	\label{fig:perf_ablation}
	\Description{The bar chart compares the relative performance improvements over SMAC for different ablated versions of Centrum across 4 benchmarks. It shows that both the co-training component and the conformal component of Centrum are necessary for achieving the best performance gain.}
\end{figure}

We conduct an ablation for the primary algorithm components of \project, which are the SGBE (Stochastic Gradient Boosting Ensemble) base surrogate model, the generalized locally adaptive conformal inference-based uncertainty estimator, and the co-training component. First, we disable co-training in \project and denote it "w/o co-training". The "w/o co-training" uses straightforward average aggregation to form the ensemble. Second, we, in addition, disable the conformal inference in "w/o co-training" and reduce it to "w/o co-training \& conformal". "w/o co-training \& conformal" is equivalent to SGBE (Stochastic Gradient Boosting), i.e., bootstrapped stochastic gradient boosting machines. \Cref{fig:perf_ablation} shows performance improvements made by \project and its two reduced variants over SMAC. On average, conformal inference contributes to about half (47.14\%) of the \project's improvement over SMAC; while co-training and SGBE base surrogate model contribute 24.98\% and 27.88\%.
The result further asserts the importance of uncertainty estimation for model-based DBMS auto-tuners.

\section{Related Work}
A complete database configuration tuning system may include several components \cite{zhao2023automatic}, as depicted below.
(a) \textbf{Core optimizer and knob selection.} A comprehensive experimental study \cite{Bin2022} for database configuration tuning
investigates different hyper-parameter optimization algorithms (e.g., BOs, reinforcement learning, genetic algorithm) and various importance measurements (e.g., Lasso, fANOVA, SHAP) in knob selection from a broader view of the machine learning community.  
(b) \textbf{Knowledge transfer for workload variation.}  OtterTune \cite{van2017automatic} transfers trained surrogate models from previous workloads to similar unseen ones by matching workload fingerprints. CGPTuner \cite{cereda2021cgptuner}, Tuneful \cite{fekry2020tune}, ResTune \cite{zhang2021restune} and ONLINE-TUNE \cite{zhang2022towards} use GP-based contextual bandit, incremental BO, meta-learning and contextual BO to adapt to workload variation, respectively. 
(c) \textbf{Human in the loop.} RelM \cite{kunjir2020black} tunes memory-based analytics and utilizes white-box expert rules to guide tuning. 
(d) \textbf{Performance \& Resource tuning.} ResTune \cite{zhang2021restune} uses constrained BO to automatically optimize resource utilization by tuning DBMS knobs without violating SLAs. 
(e) \textbf{Configuration safety assessment.} ONLINE-TUNE \cite{zhang2022towards} leverages black-box and white-box knowledge to build a evaluate the safety of configurations and avoid tuning-incurred DBMS collapses. 
Core optimizer, knob selection, knowledge transfer, expert and white-box knowledge, performance \& resource-efficiency co-tuning, and finally guaranteeing configuration safety, jointly make a global technical roadmap to build contemporary DBMS auto-tuners.
In this paper, we focus on tuning the DBMS performance via an enhanced core optimizer with a distribution-free tree-ensemble-based surrogate model.

\section{Conclusion}
\label{sec:conclusion}
We quantitatively analyze the limitations of Gaussian process-based Bayesian optimization in real-world DBMS performance tuning and re-design a model-based DBMS auto-tuner with minimal distributional assumptions.
We propose a new gradient boosting ensemble model-based framework, \project, which systematically lifts the compromised assumptions of Gaussian processes in surrogate modeling and exhibits superior optimizability toward DBMS performance. \project features modern statistical learning techniques that include stochastic gradient boosting ensembles for point prediction and locally adaptive conformal inference for interval estimation, and further boost their accuracy via a surrogate fine-tuning strategy to realize optimal ensembles.
Comprehensive experiments show \project improves beyond existing DBMS auto-tuners with better-tuned performance and less trial-and-error cost.

\bibliographystyle{ACM-Reference-Format}
\bibliography{centrum_ref_clean}


\begin{thebibliography}{73}


\ifx \showCODEN    \undefined \def \showCODEN     #1{\unskip}     \fi
\ifx \showDOI      \undefined \def \showDOI       #1{#1}\fi
\ifx \showISBNx    \undefined \def \showISBNx     #1{\unskip}     \fi
\ifx \showISBNxiii \undefined \def \showISBNxiii  #1{\unskip}     \fi
\ifx \showISSN     \undefined \def \showISSN      #1{\unskip}     \fi
\ifx \showLCCN     \undefined \def \showLCCN      #1{\unskip}     \fi
\ifx \shownote     \undefined \def \shownote      #1{#1}          \fi
\ifx \showarticletitle \undefined \def \showarticletitle #1{#1}   \fi
\ifx \showURL      \undefined \def \showURL       {\relax}        \fi
\providecommand\bibfield[2]{#2}
\providecommand\bibinfo[2]{#2}
\providecommand\natexlab[1]{#1}
\providecommand\showeprint[2][]{arXiv:#2}

\bibitem[wie(2022)]%
        {wiebe2022robust}
 \bibinfo{year}{2022}\natexlab{}.
\newblock \showarticletitle{A robust approach to warped Gaussian
  process-constrained optimization}.
\newblock \bibinfo{journal}{\emph{Mathematical Programming}}
  \bibinfo{volume}{196}, \bibinfo{number}{1} (\bibinfo{year}{2022}),
  \bibinfo{pages}{805--839}.
\newblock


\bibitem[Ament et~al\mbox{.}(2023)]%
        {ament2023unexpected}
\bibfield{author}{\bibinfo{person}{Sebastian Ament}, \bibinfo{person}{Samuel
  Daulton}, \bibinfo{person}{David Eriksson}, \bibinfo{person}{Maximilian
  Balandat}, {and} \bibinfo{person}{Eytan Bakshy}.}
  \bibinfo{year}{2023}\natexlab{}.
\newblock \showarticletitle{Unexpected improvements to expected improvement for
  bayesian optimization}.
\newblock \bibinfo{journal}{\emph{Advances in Neural Information Processing
  Systems}}  \bibinfo{volume}{36} (\bibinfo{year}{2023}),
  \bibinfo{pages}{20577--20612}.
\newblock


\bibitem[Angelopoulos and Bates(2021)]%
        {angelopoulos2021gentle}
\bibfield{author}{\bibinfo{person}{Anastasios~N Angelopoulos} {and}
  \bibinfo{person}{Stephen Bates}.} \bibinfo{year}{2021}\natexlab{}.
\newblock \showarticletitle{A gentle introduction to conformal prediction and
  distribution-free uncertainty quantification}.
\newblock \bibinfo{journal}{\emph{arXiv preprint arXiv:2107.07511}}
  (\bibinfo{year}{2021}).
\newblock


\bibitem[Borisov et~al\mbox{.}(2022)]%
        {borisov2022deep}
\bibfield{author}{\bibinfo{person}{Vadim Borisov}, \bibinfo{person}{Tobias
  Leemann}, \bibinfo{person}{Kathrin Se{\ss}ler}, \bibinfo{person}{Johannes
  Haug}, \bibinfo{person}{Martin Pawelczyk}, {and} \bibinfo{person}{Gjergji
  Kasneci}.} \bibinfo{year}{2022}\natexlab{}.
\newblock \showarticletitle{Deep neural networks and tabular data: A survey}.
\newblock \bibinfo{journal}{\emph{IEEE transactions on neural networks and
  learning systems}} (\bibinfo{year}{2022}).
\newblock


\bibitem[Breiman(1996)]%
        {breiman1996bagging}
\bibfield{author}{\bibinfo{person}{Leo Breiman}.}
  \bibinfo{year}{1996}\natexlab{}.
\newblock \showarticletitle{Bagging predictors}.
\newblock \bibinfo{journal}{\emph{Machine learning}}  \bibinfo{volume}{24}
  (\bibinfo{year}{1996}), \bibinfo{pages}{123--140}.
\newblock


\bibitem[Breiman(2001)]%
        {breiman2001random}
\bibfield{author}{\bibinfo{person}{Leo Breiman}.}
  \bibinfo{year}{2001}\natexlab{}.
\newblock \showarticletitle{Random forests}.
\newblock \bibinfo{journal}{\emph{Machine learning}}  \bibinfo{volume}{45}
  (\bibinfo{year}{2001}), \bibinfo{pages}{5--32}.
\newblock


\bibitem[Cereda et~al\mbox{.}(2021)]%
        {cereda2021cgptuner}
\bibfield{author}{\bibinfo{person}{Stefano Cereda}, \bibinfo{person}{Stefano
  Valladares}, \bibinfo{person}{Paolo Cremonesi}, {and}
  \bibinfo{person}{Stefano Doni}.} \bibinfo{year}{2021}\natexlab{}.
\newblock \showarticletitle{Cgptuner: a contextual gaussian process bandit
  approach for the automatic tuning of it configurations under varying workload
  conditions}.
\newblock \bibinfo{journal}{\emph{Proceedings of the VLDB Endowment}}
  \bibinfo{volume}{14}, \bibinfo{number}{8} (\bibinfo{year}{2021}),
  \bibinfo{pages}{1401--1413}.
\newblock


\bibitem[Chen and Guestrin(2016)]%
        {chen2016xgboost}
\bibfield{author}{\bibinfo{person}{Tianqi Chen} {and} \bibinfo{person}{Carlos
  Guestrin}.} \bibinfo{year}{2016}\natexlab{}.
\newblock \showarticletitle{Xgboost: A scalable tree boosting system}. In
  \bibinfo{booktitle}{\emph{Proceedings of the 22nd acm sigkdd international
  conference on knowledge discovery and data mining}}.
  \bibinfo{pages}{785--794}.
\newblock


\bibitem[Colombo(2023)]%
        {colombo2023training}
\bibfield{author}{\bibinfo{person}{Nicolo Colombo}.}
  \bibinfo{year}{2023}\natexlab{}.
\newblock \showarticletitle{On training locally adaptive CP}. In
  \bibinfo{booktitle}{\emph{Conformal and Probabilistic Prediction with
  Applications}}. PMLR, \bibinfo{pages}{384--398}.
\newblock


\bibitem[Cowen-Rivers et~al\mbox{.}(2022)]%
        {HEBO}
\bibfield{author}{\bibinfo{person}{Alexander~I. Cowen-Rivers},
  \bibinfo{person}{Wenlong Lyu}, \bibinfo{person}{Rasul Tutunov},
  \bibinfo{person}{Zhi Wang}, \bibinfo{person}{Antoine Grosnit},
  \bibinfo{person}{Ryan~Rhys Griffiths}, \bibinfo{person}{Alexandre~Max
  Maraval}, \bibinfo{person}{Hao Jianye}, \bibinfo{person}{Jun Wang},
  \bibinfo{person}{Jan Peters}, {and} \bibinfo{person}{Haitham Bou-Ammar}.}
  \bibinfo{year}{2022}\natexlab{}.
\newblock \showarticletitle{HEBO: Pushing The Limits of Sample-Efficient
  Hyper-parameter Optimisation}.
\newblock \bibinfo{journal}{\emph{J. Artif. Int. Res.}}  \bibinfo{volume}{74}
  (\bibinfo{date}{sep} \bibinfo{year}{2022}), \bibinfo{numpages}{81}~pages.
\newblock
\showISSN{1076-9757}
\urldef\tempurl%
\url{https://doi.org/10.1613/jair.1.13643}
\showDOI{\tempurl}


\bibitem[Curino et~al\mbox{.}(2020)]%
        {curino2020mlos}
\bibfield{author}{\bibinfo{person}{Carlo Curino}, \bibinfo{person}{Neha
  Godwal}, \bibinfo{person}{Brian Kroth}, \bibinfo{person}{Sergiy Kuryata},
  \bibinfo{person}{Greg Lapinski}, \bibinfo{person}{Siqi Liu},
  \bibinfo{person}{Slava Oks}, \bibinfo{person}{Olga Poppe},
  \bibinfo{person}{Adam Smiechowski}, \bibinfo{person}{Ed Thayer},
  {et~al\mbox{.}}} \bibinfo{year}{2020}\natexlab{}.
\newblock \showarticletitle{MLOS: An infrastructure for automated software
  performance engineering}. In \bibinfo{booktitle}{\emph{Proceedings of the
  Fourth International Workshop on Data Management for End-to-End Machine
  Learning}}. \bibinfo{pages}{1--5}.
\newblock


\bibitem[Denuit et~al\mbox{.}(2006)]%
        {denuit2006actuarial}
\bibfield{author}{\bibinfo{person}{Michel Denuit}, \bibinfo{person}{Jan
  Dhaene}, \bibinfo{person}{Marc Goovaerts}, {and} \bibinfo{person}{Rob Kaas}.}
  \bibinfo{year}{2006}\natexlab{}.
\newblock \bibinfo{booktitle}{\emph{Actuarial theory for dependent risks:
  measures, orders and models}}.
\newblock \bibinfo{publisher}{John Wiley \& Sons}.
\newblock


\bibitem[Deshwal et~al\mbox{.}(2021)]%
        {deshwal2021bayesian}
\bibfield{author}{\bibinfo{person}{Aryan Deshwal}, \bibinfo{person}{Syrine
  Belakaria}, {and} \bibinfo{person}{Janardhan~Rao Doppa}.}
  \bibinfo{year}{2021}\natexlab{}.
\newblock \showarticletitle{Bayesian optimization over hybrid spaces}. In
  \bibinfo{booktitle}{\emph{International Conference on Machine Learning}}.
  PMLR, \bibinfo{pages}{2632--2643}.
\newblock


\bibitem[Duan et~al\mbox{.}(2009)]%
        {duan2009tuning}
\bibfield{author}{\bibinfo{person}{Songyun Duan}, \bibinfo{person}{Vamsidhar
  Thummala}, {and} \bibinfo{person}{Shivnath Babu}.}
  \bibinfo{year}{2009}\natexlab{}.
\newblock \showarticletitle{Tuning database configuration parameters with
  ituned}.
\newblock \bibinfo{journal}{\emph{Proceedings of the VLDB Endowment}}
  \bibinfo{volume}{2}, \bibinfo{number}{1} (\bibinfo{year}{2009}),
  \bibinfo{pages}{1246--1257}.
\newblock


\bibitem[Duan et~al\mbox{.}(2020)]%
        {duan2020ngboost}
\bibfield{author}{\bibinfo{person}{Tony Duan}, \bibinfo{person}{Avati Anand},
  \bibinfo{person}{Daisy~Yi Ding}, \bibinfo{person}{Khanh~K Thai},
  \bibinfo{person}{Sanjay Basu}, \bibinfo{person}{Andrew Ng}, {and}
  \bibinfo{person}{Alejandro Schuler}.} \bibinfo{year}{2020}\natexlab{}.
\newblock \showarticletitle{Ngboost: Natural gradient boosting for
  probabilistic prediction}. In \bibinfo{booktitle}{\emph{International
  conference on machine learning}}. PMLR, \bibinfo{pages}{2690--2700}.
\newblock


\bibitem[Eriksson et~al\mbox{.}(2019)]%
        {eriksson2019scalable}
\bibfield{author}{\bibinfo{person}{David Eriksson}, \bibinfo{person}{Michael
  Pearce}, \bibinfo{person}{Jacob Gardner}, \bibinfo{person}{Ryan~D Turner},
  {and} \bibinfo{person}{Matthias Poloczek}.} \bibinfo{year}{2019}\natexlab{}.
\newblock \showarticletitle{Scalable global optimization via local Bayesian
  optimization}.
\newblock \bibinfo{journal}{\emph{Advances in neural information processing
  systems}}  \bibinfo{volume}{32} (\bibinfo{year}{2019}).
\newblock


\bibitem[Fekry et~al\mbox{.}(2020)]%
        {fekry2020tune}
\bibfield{author}{\bibinfo{person}{Ayat Fekry}, \bibinfo{person}{Lucian
  Carata}, \bibinfo{person}{Thomas Pasquier}, \bibinfo{person}{Andrew Rice},
  {and} \bibinfo{person}{Andy Hopper}.} \bibinfo{year}{2020}\natexlab{}.
\newblock \showarticletitle{To tune or not to tune? in search of optimal
  configurations for data analytics}. In \bibinfo{booktitle}{\emph{Proceedings
  of the 26th ACM SIGKDD International Conference on Knowledge Discovery \&
  Data Mining}}. \bibinfo{pages}{2494--2504}.
\newblock


\bibitem[Frazier(2018)]%
        {frazier2018tutorial}
\bibfield{author}{\bibinfo{person}{Peter~I Frazier}.}
  \bibinfo{year}{2018}\natexlab{}.
\newblock \showarticletitle{A tutorial on Bayesian optimization}.
\newblock \bibinfo{journal}{\emph{arXiv preprint arXiv:1807.02811}}
  (\bibinfo{year}{2018}).
\newblock


\bibitem[Friedman(2001)]%
        {friedman2001greedy}
\bibfield{author}{\bibinfo{person}{Jerome~H Friedman}.}
  \bibinfo{year}{2001}\natexlab{}.
\newblock \showarticletitle{Greedy function approximation: a gradient boosting
  machine}.
\newblock \bibinfo{journal}{\emph{Annals of statistics}}
  (\bibinfo{year}{2001}), \bibinfo{pages}{1189--1232}.
\newblock


\bibitem[Friedman(2002)]%
        {friedman2002stochastic}
\bibfield{author}{\bibinfo{person}{Jerome~H Friedman}.}
  \bibinfo{year}{2002}\natexlab{}.
\newblock \showarticletitle{Stochastic gradient boosting}.
\newblock \bibinfo{journal}{\emph{Computational statistics \& data analysis}}
  \bibinfo{volume}{38}, \bibinfo{number}{4} (\bibinfo{year}{2002}),
  \bibinfo{pages}{367--378}.
\newblock


\bibitem[Gal et~al\mbox{.}(2016)]%
        {gal2016uncertainty}
\bibfield{author}{\bibinfo{person}{Yarin Gal} {et~al\mbox{.}}}
  \bibinfo{year}{2016}\natexlab{}.
\newblock \showarticletitle{Uncertainty in deep learning}.
\newblock  (\bibinfo{year}{2016}).
\newblock


\bibitem[Gorishniy et~al\mbox{.}(2021)]%
        {gorishniy2021revisiting}
\bibfield{author}{\bibinfo{person}{Yury Gorishniy}, \bibinfo{person}{Ivan
  Rubachev}, \bibinfo{person}{Valentin Khrulkov}, {and} \bibinfo{person}{Artem
  Babenko}.} \bibinfo{year}{2021}\natexlab{}.
\newblock \showarticletitle{Revisiting deep learning models for tabular data}.
\newblock \bibinfo{journal}{\emph{Advances in Neural Information Processing
  Systems}}  \bibinfo{volume}{34} (\bibinfo{year}{2021}),
  \bibinfo{pages}{18932--18943}.
\newblock


\bibitem[Grinsztajn et~al\mbox{.}(2022)]%
        {grinsztajn2022tree}
\bibfield{author}{\bibinfo{person}{L{\'e}o Grinsztajn},
  \bibinfo{person}{Edouard Oyallon}, {and} \bibinfo{person}{Ga{\"e}l
  Varoquaux}.} \bibinfo{year}{2022}\natexlab{}.
\newblock \showarticletitle{Why do tree-based models still outperform deep
  learning on typical tabular data?}
\newblock \bibinfo{journal}{\emph{Advances in neural information processing
  systems}}  \bibinfo{volume}{35} (\bibinfo{year}{2022}),
  \bibinfo{pages}{507--520}.
\newblock


\bibitem[Gupta et~al\mbox{.}(2022)]%
        {gupta2022nested}
\bibfield{author}{\bibinfo{person}{Chirag Gupta}, \bibinfo{person}{Arun~K
  Kuchibhotla}, {and} \bibinfo{person}{Aaditya Ramdas}.}
  \bibinfo{year}{2022}\natexlab{}.
\newblock \showarticletitle{Nested conformal prediction and quantile out-of-bag
  ensemble methods}.
\newblock \bibinfo{journal}{\emph{Pattern Recognition}}  \bibinfo{volume}{127}
  (\bibinfo{year}{2022}), \bibinfo{pages}{108496}.
\newblock


\bibitem[Head et~al\mbox{.}(2021)]%
        {skopt}
\bibfield{author}{\bibinfo{person}{Tim Head}, \bibinfo{person}{Manoj Kumar},
  \bibinfo{person}{Holger Nahrstaedt}, \bibinfo{person}{Gilles Louppe}, {and}
  \bibinfo{person}{Iarosla Shcherbatyi}.} \bibinfo{year}{2021}\natexlab{}.
\newblock \bibinfo{booktitle}{\emph{scikit-optimize/scikit-optimize}}.
\newblock
\urldef\tempurl%
\url{https://doi.org/10.5281/zenodo.5565057}
\showDOI{\tempurl}


\bibitem[Hutter et~al\mbox{.}(2011)]%
        {hutter2011}
\bibfield{author}{\bibinfo{person}{Frank Hutter}, \bibinfo{person}{Holger~H
  Hoos}, {and} \bibinfo{person}{Kevin Leyton-Brown}.}
  \bibinfo{year}{2011}\natexlab{}.
\newblock \showarticletitle{Sequential model-based optimization for general
  algorithm configuration}. In \bibinfo{booktitle}{\emph{Learning and
  Intelligent Optimization: 5th International Conference, LION 5, Rome, Italy,
  January 17-21, 2011. Selected Papers 5}}. Springer,
  \bibinfo{pages}{507--523}.
\newblock


\bibitem[Jiang et~al\mbox{.}(2024)]%
        {JMLR:v25:23-0537}
\bibfield{author}{\bibinfo{person}{Huaijun Jiang}, \bibinfo{person}{Yu Shen},
  \bibinfo{person}{Yang Li}, \bibinfo{person}{Beicheng Xu},
  \bibinfo{person}{Sixian Du}, \bibinfo{person}{Wentao Zhang},
  \bibinfo{person}{Ce Zhang}, {and} \bibinfo{person}{Bin Cui}.}
  \bibinfo{year}{2024}\natexlab{}.
\newblock \showarticletitle{OpenBox: A Python Toolkit for Generalized Black-box
  Optimization}.
\newblock \bibinfo{journal}{\emph{Journal of Machine Learning Research}}
  \bibinfo{volume}{25}, \bibinfo{number}{120} (\bibinfo{year}{2024}),
  \bibinfo{pages}{1--11}.
\newblock
\urldef\tempurl%
\url{http://jmlr.org/papers/v25/23-0537.html}
\showURL{%
\tempurl}


\bibitem[Johansson et~al\mbox{.}(2014)]%
        {johansson2014regression}
\bibfield{author}{\bibinfo{person}{Ulf Johansson}, \bibinfo{person}{Henrik
  Bostr{\"o}m}, \bibinfo{person}{Tuve L{\"o}fstr{\"o}m}, {and}
  \bibinfo{person}{Henrik Linusson}.} \bibinfo{year}{2014}\natexlab{}.
\newblock \showarticletitle{Regression conformal prediction with random
  forests}.
\newblock \bibinfo{journal}{\emph{Machine learning}}  \bibinfo{volume}{97}
  (\bibinfo{year}{2014}), \bibinfo{pages}{155--176}.
\newblock


\bibitem[Jones et~al\mbox{.}(1998)]%
        {jones1998efficient}
\bibfield{author}{\bibinfo{person}{Donald~R Jones}, \bibinfo{person}{Matthias
  Schonlau}, {and} \bibinfo{person}{William~J Welch}.}
  \bibinfo{year}{1998}\natexlab{}.
\newblock \showarticletitle{Efficient global optimization of expensive
  black-box functions}.
\newblock \bibinfo{journal}{\emph{Journal of Global optimization}}
  \bibinfo{volume}{13} (\bibinfo{year}{1998}), \bibinfo{pages}{455--492}.
\newblock


\bibitem[Kanellis et~al\mbox{.}(2022)]%
        {kanellis2022llamatune}
\bibfield{author}{\bibinfo{person}{Konstantinos Kanellis},
  \bibinfo{person}{Cong Ding}, \bibinfo{person}{Brian Kroth},
  \bibinfo{person}{Andreas M{\"u}ller}, \bibinfo{person}{Carlo Curino}, {and}
  \bibinfo{person}{Shivaram Venkataraman}.} \bibinfo{year}{2022}\natexlab{}.
\newblock \showarticletitle{LlamaTune: sample-efficient DBMS configuration
  tuning}.
\newblock \bibinfo{journal}{\emph{arXiv preprint arXiv:2203.05128}}
  (\bibinfo{year}{2022}).
\newblock


\bibitem[Ke et~al\mbox{.}(2017)]%
        {ke2017lightgbm}
\bibfield{author}{\bibinfo{person}{Guolin Ke}, \bibinfo{person}{Qi Meng},
  \bibinfo{person}{Thomas Finley}, \bibinfo{person}{Taifeng Wang},
  \bibinfo{person}{Wei Chen}, \bibinfo{person}{Weidong Ma},
  \bibinfo{person}{Qiwei Ye}, {and} \bibinfo{person}{Tie-Yan Liu}.}
  \bibinfo{year}{2017}\natexlab{}.
\newblock \showarticletitle{Lightgbm: A highly efficient gradient boosting
  decision tree}.
\newblock \bibinfo{journal}{\emph{Advances in neural information processing
  systems}}  \bibinfo{volume}{30} (\bibinfo{year}{2017}).
\newblock


\bibitem[Kim et~al\mbox{.}(2020)]%
        {kim2020predictive}
\bibfield{author}{\bibinfo{person}{Byol Kim}, \bibinfo{person}{Chen Xu}, {and}
  \bibinfo{person}{Rina Barber}.} \bibinfo{year}{2020}\natexlab{}.
\newblock \showarticletitle{Predictive inference is free with the
  jackknife+-after-bootstrap}.
\newblock \bibinfo{journal}{\emph{Advances in Neural Information Processing
  Systems}}  \bibinfo{volume}{33} (\bibinfo{year}{2020}),
  \bibinfo{pages}{4138--4149}.
\newblock


\bibitem[Kunjir and Babu(2020)]%
        {kunjir2020black}
\bibfield{author}{\bibinfo{person}{Mayuresh Kunjir} {and}
  \bibinfo{person}{Shivnath Babu}.} \bibinfo{year}{2020}\natexlab{}.
\newblock \showarticletitle{Black or white? how to develop an autotuner for
  memory-based analytics}. In \bibinfo{booktitle}{\emph{Proceedings of the 2020
  ACM SIGMOD International Conference on Management of Data}}.
  \bibinfo{pages}{1667--1683}.
\newblock


\bibitem[Lakshminarayanan et~al\mbox{.}(2017)]%
        {lakshminarayanan2017simple}
\bibfield{author}{\bibinfo{person}{Balaji Lakshminarayanan},
  \bibinfo{person}{Alexander Pritzel}, {and} \bibinfo{person}{Charles
  Blundell}.} \bibinfo{year}{2017}\natexlab{}.
\newblock \showarticletitle{Simple and scalable predictive uncertainty
  estimation using deep ensembles}.
\newblock \bibinfo{journal}{\emph{Advances in neural information processing
  systems}}  \bibinfo{volume}{30} (\bibinfo{year}{2017}).
\newblock


\bibitem[Lei et~al\mbox{.}(2018)]%
        {lei2018distribution}
\bibfield{author}{\bibinfo{person}{Jing Lei}, \bibinfo{person}{Max G’Sell},
  \bibinfo{person}{Alessandro Rinaldo}, \bibinfo{person}{Ryan~J Tibshirani},
  {and} \bibinfo{person}{Larry Wasserman}.} \bibinfo{year}{2018}\natexlab{}.
\newblock \showarticletitle{Distribution-free predictive inference for
  regression}.
\newblock \bibinfo{journal}{\emph{J. Amer. Statist. Assoc.}}
  \bibinfo{volume}{113}, \bibinfo{number}{523} (\bibinfo{year}{2018}),
  \bibinfo{pages}{1094--1111}.
\newblock


\bibitem[Lessmann et~al\mbox{.}(2005)]%
        {GA2005}
\bibfield{author}{\bibinfo{person}{Stefan Lessmann}, \bibinfo{person}{Robert
  Stahlbock}, {and} \bibinfo{person}{Sven~F Crone}.}
  \bibinfo{year}{2005}\natexlab{}.
\newblock \showarticletitle{Optimizing hyperparameters of support vector
  machines by genetic algorithms.}. In \bibinfo{booktitle}{\emph{IC-AI}},
  Vol.~\bibinfo{volume}{74}. \bibinfo{pages}{82}.
\newblock


\bibitem[Li et~al\mbox{.}(2019)]%
        {li2019qtune}
\bibfield{author}{\bibinfo{person}{Guoliang Li}, \bibinfo{person}{Xuanhe Zhou},
  \bibinfo{person}{Shifu Li}, {and} \bibinfo{person}{Bo Gao}.}
  \bibinfo{year}{2019}\natexlab{}.
\newblock \showarticletitle{Qtune: A query-aware database tuning system with
  deep reinforcement learning}.
\newblock \bibinfo{journal}{\emph{Proceedings of the VLDB Endowment}}
  \bibinfo{volume}{12}, \bibinfo{number}{12} (\bibinfo{year}{2019}),
  \bibinfo{pages}{2118--2130}.
\newblock


\bibitem[Li et~al\mbox{.}(2021)]%
        {li2021openbox}
\bibfield{author}{\bibinfo{person}{Yang Li}, \bibinfo{person}{Yu Shen},
  \bibinfo{person}{Wentao Zhang}, \bibinfo{person}{Yuanwei Chen},
  \bibinfo{person}{Huaijun Jiang}, \bibinfo{person}{Mingchao Liu},
  \bibinfo{person}{Jiawei Jiang}, \bibinfo{person}{Jinyang Gao},
  \bibinfo{person}{Wentao Wu}, \bibinfo{person}{Zhi Yang}, {et~al\mbox{.}}}
  \bibinfo{year}{2021}\natexlab{}.
\newblock \showarticletitle{Openbox: A generalized black-box optimization
  service}. In \bibinfo{booktitle}{\emph{Proceedings of the 27th ACM SIGKDD
  Conference on Knowledge Discovery \& Data Mining}}.
  \bibinfo{pages}{3209--3219}.
\newblock


\bibitem[Lindauer et~al\mbox{.}(2022)]%
        {lindauer2022smac3}
\bibfield{author}{\bibinfo{person}{Marius Lindauer}, \bibinfo{person}{Katharina
  Eggensperger}, \bibinfo{person}{Matthias Feurer}, \bibinfo{person}{Andr{\'e}
  Biedenkapp}, \bibinfo{person}{Difan Deng}, \bibinfo{person}{Carolin
  Benjamins}, \bibinfo{person}{Tim Ruhkopf}, \bibinfo{person}{Ren{\'e} Sass},
  {and} \bibinfo{person}{Frank Hutter}.} \bibinfo{year}{2022}\natexlab{}.
\newblock \showarticletitle{SMAC3: A versatile Bayesian optimization package
  for hyperparameter optimization}.
\newblock \bibinfo{journal}{\emph{The Journal of Machine Learning Research}}
  \bibinfo{volume}{23}, \bibinfo{number}{1} (\bibinfo{year}{2022}),
  \bibinfo{pages}{2475--2483}.
\newblock


\bibitem[Linusson et~al\mbox{.}(2020)]%
        {linusson2020efficient}
\bibfield{author}{\bibinfo{person}{Henrik Linusson}, \bibinfo{person}{Ulf
  Johansson}, {and} \bibinfo{person}{Henrik Bostr{\"o}m}.}
  \bibinfo{year}{2020}\natexlab{}.
\newblock \showarticletitle{Efficient conformal predictor ensembles}.
\newblock \bibinfo{journal}{\emph{Neurocomputing}}  \bibinfo{volume}{397}
  (\bibinfo{year}{2020}), \bibinfo{pages}{266--278}.
\newblock


\bibitem[Makridakis et~al\mbox{.}(2022)]%
        {makridakis2022m5}
\bibfield{author}{\bibinfo{person}{Spyros Makridakis},
  \bibinfo{person}{Evangelos Spiliotis}, {and} \bibinfo{person}{Vassilios
  Assimakopoulos}.} \bibinfo{year}{2022}\natexlab{}.
\newblock \showarticletitle{M5 accuracy competition: Results, findings, and
  conclusions}.
\newblock \bibinfo{journal}{\emph{International Journal of Forecasting}}
  \bibinfo{volume}{38}, \bibinfo{number}{4} (\bibinfo{year}{2022}),
  \bibinfo{pages}{1346--1364}.
\newblock


\bibitem[Malinin et~al\mbox{.}(2021)]%
        {malinin2021uncertainty}
\bibfield{author}{\bibinfo{person}{Andrey Malinin}, \bibinfo{person}{Liudmila
  Prokhorenkova}, {and} \bibinfo{person}{Aleksei Ustimenko}.}
  \bibinfo{year}{2021}\natexlab{}.
\newblock \showarticletitle{Uncertainty in Gradient Boosting via Ensembles}. In
  \bibinfo{booktitle}{\emph{International Conference on Learning
  Representations}}.
\newblock
\urldef\tempurl%
\url{https://openreview.net/forum?id=1Jv6b0Zq3qi}
\showURL{%
\tempurl}


\bibitem[McElfresh et~al\mbox{.}(2024)]%
        {mcelfresh2024neural}
\bibfield{author}{\bibinfo{person}{Duncan McElfresh}, \bibinfo{person}{Sujay
  Khandagale}, \bibinfo{person}{Jonathan Valverde}, \bibinfo{person}{Vishak
  Prasad~C}, \bibinfo{person}{Ganesh Ramakrishnan}, \bibinfo{person}{Micah
  Goldblum}, {and} \bibinfo{person}{Colin White}.}
  \bibinfo{year}{2024}\natexlab{}.
\newblock \showarticletitle{When do neural nets outperform boosted trees on
  tabular data?}
\newblock \bibinfo{journal}{\emph{Advances in Neural Information Processing
  Systems}}  \bibinfo{volume}{36} (\bibinfo{year}{2024}).
\newblock


\bibitem[Papadopoulos et~al\mbox{.}(2008)]%
        {papadopoulos2008normalized}
\bibfield{author}{\bibinfo{person}{Harris Papadopoulos}, \bibinfo{person}{Alex
  Gammerman}, {and} \bibinfo{person}{Volodya Vovk}.}
  \bibinfo{year}{2008}\natexlab{}.
\newblock \showarticletitle{Normalized nonconformity measures for regression
  conformal prediction}. In \bibinfo{booktitle}{\emph{Proceedings of the IASTED
  International Conference on Artificial Intelligence and Applications (AIA
  2008)}}. \bibinfo{pages}{64--69}.
\newblock


\bibitem[Paszke et~al\mbox{.}(2019)]%
        {NEURIPS2019_9015}
\bibfield{author}{\bibinfo{person}{Adam Paszke}, \bibinfo{person}{Sam Gross},
  \bibinfo{person}{Francisco Massa}, \bibinfo{person}{Adam Lerer},
  \bibinfo{person}{James Bradbury}, \bibinfo{person}{Gregory Chanan},
  \bibinfo{person}{Trevor Killeen}, \bibinfo{person}{Zeming Lin},
  \bibinfo{person}{Natalia Gimelshein}, \bibinfo{person}{Luca Antiga},
  \bibinfo{person}{Alban Desmaison}, \bibinfo{person}{Andreas Kopf},
  \bibinfo{person}{Edward Yang}, \bibinfo{person}{Zachary DeVito},
  \bibinfo{person}{Martin Raison}, \bibinfo{person}{Alykhan Tejani},
  \bibinfo{person}{Sasank Chilamkurthy}, \bibinfo{person}{Benoit Steiner},
  \bibinfo{person}{Lu Fang}, \bibinfo{person}{Junjie Bai}, {and}
  \bibinfo{person}{Soumith Chintala}.} \bibinfo{year}{2019}\natexlab{}.
\newblock \showarticletitle{PyTorch: An Imperative Style, High-Performance Deep
  Learning Library}.
\newblock In \bibinfo{booktitle}{\emph{Advances in Neural Information
  Processing Systems 32}}. \bibinfo{publisher}{Curran Associates, Inc.},
  \bibinfo{pages}{8024--8035}.
\newblock
\urldef\tempurl%
\url{http://papers.neurips.cc/paper/9015-pytorch-an-imperative-style-high-performance-deep-learning-library.pdf}
\showURL{%
\tempurl}


\bibitem[Pedregosa et~al\mbox{.}(2011)]%
        {scikit-learn}
\bibfield{author}{\bibinfo{person}{F. Pedregosa}, \bibinfo{person}{G.
  Varoquaux}, \bibinfo{person}{A. Gramfort}, \bibinfo{person}{V. Michel},
  \bibinfo{person}{B. Thirion}, \bibinfo{person}{O. Grisel},
  \bibinfo{person}{M. Blondel}, \bibinfo{person}{P. Prettenhofer},
  \bibinfo{person}{R. Weiss}, \bibinfo{person}{V. Dubourg}, \bibinfo{person}{J.
  Vanderplas}, \bibinfo{person}{A. Passos}, \bibinfo{person}{D. Cournapeau},
  \bibinfo{person}{M. Brucher}, \bibinfo{person}{M. Perrot}, {and}
  \bibinfo{person}{E. Duchesnay}.} \bibinfo{year}{2011}\natexlab{}.
\newblock \showarticletitle{Scikit-learn: Machine Learning in {P}ython}.
\newblock \bibinfo{journal}{\emph{Journal of Machine Learning Research}}
  \bibinfo{volume}{12} (\bibinfo{year}{2011}), \bibinfo{pages}{2825--2830}.
\newblock


\bibitem[Rubinstein and Kroese(2004)]%
        {rubinstein2004cross}
\bibfield{author}{\bibinfo{person}{Reuven~Y Rubinstein} {and}
  \bibinfo{person}{Dirk~P Kroese}.} \bibinfo{year}{2004}\natexlab{}.
\newblock \bibinfo{booktitle}{\emph{The cross-entropy method: a unified
  approach to combinatorial optimization, Monte-Carlo simulation, and machine
  learning}}. Vol.~\bibinfo{volume}{133}.
\newblock \bibinfo{publisher}{Springer}.
\newblock


\bibitem[Seeger(2004)]%
        {seeger2004gaussian}
\bibfield{author}{\bibinfo{person}{Matthias Seeger}.}
  \bibinfo{year}{2004}\natexlab{}.
\newblock \showarticletitle{Gaussian processes for machine learning}.
\newblock \bibinfo{journal}{\emph{International journal of neural systems}}
  \bibinfo{volume}{14}, \bibinfo{number}{02} (\bibinfo{year}{2004}),
  \bibinfo{pages}{69--106}.
\newblock


\bibitem[Sen(1968)]%
        {sen1968estimates}
\bibfield{author}{\bibinfo{person}{Pranab~Kumar Sen}.}
  \bibinfo{year}{1968}\natexlab{}.
\newblock \showarticletitle{Estimates of the regression coefficient based on
  Kendall's tau}.
\newblock \bibinfo{journal}{\emph{Journal of the American statistical
  association}} \bibinfo{volume}{63}, \bibinfo{number}{324}
  (\bibinfo{year}{1968}), \bibinfo{pages}{1379--1389}.
\newblock


\bibitem[Shafer and Vovk(2008)]%
        {shafer2008tutorial}
\bibfield{author}{\bibinfo{person}{Glenn Shafer} {and}
  \bibinfo{person}{Vladimir Vovk}.} \bibinfo{year}{2008}\natexlab{}.
\newblock \showarticletitle{A tutorial on conformal prediction.}
\newblock \bibinfo{journal}{\emph{Journal of Machine Learning Research}}
  \bibinfo{volume}{9}, \bibinfo{number}{3} (\bibinfo{year}{2008}).
\newblock


\bibitem[Shahriari et~al\mbox{.}(2015)]%
        {shahriari2015taking}
\bibfield{author}{\bibinfo{person}{Bobak Shahriari}, \bibinfo{person}{Kevin
  Swersky}, \bibinfo{person}{Ziyu Wang}, \bibinfo{person}{Ryan~P Adams}, {and}
  \bibinfo{person}{Nando De~Freitas}.} \bibinfo{year}{2015}\natexlab{}.
\newblock \showarticletitle{Taking the human out of the loop: A review of
  Bayesian optimization}.
\newblock \bibinfo{journal}{\emph{Proc. IEEE}} \bibinfo{volume}{104},
  \bibinfo{number}{1} (\bibinfo{year}{2015}), \bibinfo{pages}{148--175}.
\newblock


\bibitem[Siegrist(1997)]%
        {Random}
\bibfield{author}{\bibinfo{person}{Kyle Siegrist}.}
  \bibinfo{year}{1997}\natexlab{}.
\newblock \bibinfo{title}{Random: Probability, Mathematical Statistics,
  Stochastic Processes}.
\newblock \bibinfo{howpublished}{\url{http://www.randomservices.org/random/}}.
\newblock


\bibitem[Snoek et~al\mbox{.}(2012)]%
        {snoek2012practical}
\bibfield{author}{\bibinfo{person}{Jasper Snoek}, \bibinfo{person}{Hugo
  Larochelle}, {and} \bibinfo{person}{Ryan~P Adams}.}
  \bibinfo{year}{2012}\natexlab{}.
\newblock \showarticletitle{Practical bayesian optimization of machine learning
  algorithms}.
\newblock \bibinfo{journal}{\emph{Advances in neural information processing
  systems}}  \bibinfo{volume}{25} (\bibinfo{year}{2012}).
\newblock


\bibitem[Snoek et~al\mbox{.}(2014)]%
        {snoek2014input}
\bibfield{author}{\bibinfo{person}{Jasper Snoek}, \bibinfo{person}{Kevin
  Swersky}, \bibinfo{person}{Rich Zemel}, {and} \bibinfo{person}{Ryan Adams}.}
  \bibinfo{year}{2014}\natexlab{}.
\newblock \showarticletitle{Input warping for Bayesian optimization of
  non-stationary functions}. In \bibinfo{booktitle}{\emph{International
  conference on machine learning}}. PMLR, \bibinfo{pages}{1674--1682}.
\newblock


\bibitem[Somepalli et~al\mbox{.}(2021)]%
        {somepalli2021saint}
\bibfield{author}{\bibinfo{person}{Gowthami Somepalli}, \bibinfo{person}{Micah
  Goldblum}, \bibinfo{person}{Avi Schwarzschild}, \bibinfo{person}{C~Bayan
  Bruss}, {and} \bibinfo{person}{Tom Goldstein}.}
  \bibinfo{year}{2021}\natexlab{}.
\newblock \showarticletitle{Saint: Improved neural networks for tabular data
  via row attention and contrastive pre-training}.
\newblock \bibinfo{journal}{\emph{arXiv preprint arXiv:2106.01342}}
  (\bibinfo{year}{2021}).
\newblock


\bibitem[Song et~al\mbox{.}(2019)]%
        {song2019autoint}
\bibfield{author}{\bibinfo{person}{Weiping Song}, \bibinfo{person}{Chence Shi},
  \bibinfo{person}{Zhiping Xiao}, \bibinfo{person}{Zhijian Duan},
  \bibinfo{person}{Yewen Xu}, \bibinfo{person}{Ming Zhang}, {and}
  \bibinfo{person}{Jian Tang}.} \bibinfo{year}{2019}\natexlab{}.
\newblock \showarticletitle{Autoint: Automatic feature interaction learning via
  self-attentive neural networks}. In \bibinfo{booktitle}{\emph{Proceedings of
  the 28th ACM international conference on information and knowledge
  management}}. \bibinfo{pages}{1161--1170}.
\newblock


\bibitem[Sprangers et~al\mbox{.}(2021)]%
        {PGBM2021}
\bibfield{author}{\bibinfo{person}{Olivier Sprangers},
  \bibinfo{person}{Sebastian Schelter}, {and} \bibinfo{person}{Maarten de
  Rijke}.} \bibinfo{year}{2021}\natexlab{}.
\newblock \showarticletitle{Probabilistic Gradient Boosting Machines for
  Large-Scale Probabilistic Regression}. In
  \bibinfo{booktitle}{\emph{Proceedings of the 27th ACM SIGKDD Conference on
  Knowledge Discovery \& Data Mining}} (Virtual Event, Singapore)
  \emph{(\bibinfo{series}{KDD '21})}. \bibinfo{publisher}{Association for
  Computing Machinery}, \bibinfo{address}{New York, NY, USA},
  \bibinfo{pages}{1510--1520}.
\newblock
\showISBNx{9781450383325}
\urldef\tempurl%
\url{https://doi.org/10.1145/3447548.3467278}
\showDOI{\tempurl}


\bibitem[Srinivas et~al\mbox{.}(2009)]%
        {srinivas2009gaussian}
\bibfield{author}{\bibinfo{person}{Niranjan Srinivas}, \bibinfo{person}{Andreas
  Krause}, \bibinfo{person}{Sham~M Kakade}, {and} \bibinfo{person}{Matthias
  Seeger}.} \bibinfo{year}{2009}\natexlab{}.
\newblock \showarticletitle{Gaussian process optimization in the bandit
  setting: No regret and experimental design}.
\newblock \bibinfo{journal}{\emph{arXiv preprint arXiv:0912.3995}}
  (\bibinfo{year}{2009}).
\newblock


\bibitem[Thirumuruganathan et~al\mbox{.}(2022)]%
        {thirumuruganathan2022prediction}
\bibfield{author}{\bibinfo{person}{Saravanan Thirumuruganathan},
  \bibinfo{person}{Suraj Shetiya}, \bibinfo{person}{Nick Koudas}, {and}
  \bibinfo{person}{Gautam Das}.} \bibinfo{year}{2022}\natexlab{}.
\newblock \showarticletitle{Prediction intervals for learned cardinality
  estimation: an experimental evaluation}. In \bibinfo{booktitle}{\emph{2022
  IEEE 38th International Conference on Data Engineering (ICDE)}}. IEEE,
  \bibinfo{pages}{3051--3064}.
\newblock


\bibitem[Thornton et~al\mbox{.}(2013)]%
        {thornton2013auto}
\bibfield{author}{\bibinfo{person}{Chris Thornton}, \bibinfo{person}{Frank
  Hutter}, \bibinfo{person}{Holger~H Hoos}, {and} \bibinfo{person}{Kevin
  Leyton-Brown}.} \bibinfo{year}{2013}\natexlab{}.
\newblock \showarticletitle{Auto-WEKA: Combined selection and hyperparameter
  optimization of classification algorithms}. In
  \bibinfo{booktitle}{\emph{Proceedings of the 19th ACM SIGKDD international
  conference on Knowledge discovery and data mining}}.
  \bibinfo{pages}{847--855}.
\newblock


\bibitem[Van~Aken et~al\mbox{.}(2017)]%
        {van2017automatic}
\bibfield{author}{\bibinfo{person}{Dana Van~Aken}, \bibinfo{person}{Andrew
  Pavlo}, \bibinfo{person}{Geoffrey~J Gordon}, {and} \bibinfo{person}{Bohan
  Zhang}.} \bibinfo{year}{2017}\natexlab{}.
\newblock \showarticletitle{Automatic database management system tuning through
  large-scale machine learning}. In \bibinfo{booktitle}{\emph{Proceedings of
  the 2017 ACM international conference on management of data}}.
  \bibinfo{pages}{1009--1024}.
\newblock


\bibitem[Wang et~al\mbox{.}(2020)]%
        {wang2020learning}
\bibfield{author}{\bibinfo{person}{Linnan Wang}, \bibinfo{person}{Rodrigo
  Fonseca}, {and} \bibinfo{person}{Yuandong Tian}.}
  \bibinfo{year}{2020}\natexlab{}.
\newblock \showarticletitle{Learning search space partition for black-box
  optimization using monte carlo tree search}.
\newblock \bibinfo{journal}{\emph{Advances in Neural Information Processing
  Systems}}  \bibinfo{volume}{33} (\bibinfo{year}{2020}),
  \bibinfo{pages}{19511--19522}.
\newblock


\bibitem[Wang et~al\mbox{.}(2022)]%
        {keentune}
\bibfield{author}{\bibinfo{person}{Runzhe Wang}, \bibinfo{person}{Qinglong
  Wang}, \bibinfo{person}{Yuxi Hu}, \bibinfo{person}{Heyuan Shi},
  \bibinfo{person}{Yuheng Shen}, \bibinfo{person}{Yu Zhan},
  \bibinfo{person}{Ying Fu}, \bibinfo{person}{Zheng Liu},
  \bibinfo{person}{Xiaohai Shi}, {and} \bibinfo{person}{Yu Jiang}.}
  \bibinfo{year}{2022}\natexlab{}.
\newblock \showarticletitle{Industry practice of configuration auto-tuning for
  cloud applications and services}. In \bibinfo{booktitle}{\emph{Proceedings of
  the 30th ACM Joint European Software Engineering Conference and Symposium on
  the Foundations of Software Engineering}} \emph{(\bibinfo{series}{ESEC/FSE
  2022})}. \bibinfo{publisher}{Association for Computing Machinery},
  \bibinfo{address}{New York, NY, USA}, \bibinfo{pages}{1555--1565}.
\newblock
\showISBNx{9781450394130}
\urldef\tempurl%
\url{https://doi.org/10.1145/3540250.3558962}
\showDOI{\tempurl}


\bibitem[Wang et~al\mbox{.}(2016)]%
        {wang2016bayesian}
\bibfield{author}{\bibinfo{person}{Ziyu Wang}, \bibinfo{person}{Frank Hutter},
  \bibinfo{person}{Masrour Zoghi}, \bibinfo{person}{David Matheson}, {and}
  \bibinfo{person}{Nando De~Feitas}.} \bibinfo{year}{2016}\natexlab{}.
\newblock \showarticletitle{Bayesian optimization in a billion dimensions via
  random embeddings}.
\newblock \bibinfo{journal}{\emph{Journal of Artificial Intelligence Research}}
   \bibinfo{volume}{55} (\bibinfo{year}{2016}), \bibinfo{pages}{361--387}.
\newblock


\bibitem[Xin et~al\mbox{.}(2022)]%
        {xin2022locat}
\bibfield{author}{\bibinfo{person}{Jinhan Xin}, \bibinfo{person}{Kai Hwang},
  {and} \bibinfo{person}{Zhibin Yu}.} \bibinfo{year}{2022}\natexlab{}.
\newblock \showarticletitle{Locat: Low-overhead online configuration
  auto-tuning of spark sql applications}. In
  \bibinfo{booktitle}{\emph{Proceedings of the 2022 International Conference on
  Management of Data}}. \bibinfo{pages}{674--684}.
\newblock


\bibitem[Xu and Xie(2021)]%
        {xu2021conformal}
\bibfield{author}{\bibinfo{person}{Chen Xu} {and} \bibinfo{person}{Yao Xie}.}
  \bibinfo{year}{2021}\natexlab{}.
\newblock \showarticletitle{Conformal prediction interval for dynamic
  time-series}. In \bibinfo{booktitle}{\emph{International Conference on
  Machine Learning}}. PMLR, \bibinfo{pages}{11559--11569}.
\newblock


\bibitem[Zhang et~al\mbox{.}(2019)]%
        {zhang2019end}
\bibfield{author}{\bibinfo{person}{Ji Zhang}, \bibinfo{person}{Yu Liu},
  \bibinfo{person}{Ke Zhou}, \bibinfo{person}{Guoliang Li},
  \bibinfo{person}{Zhili Xiao}, \bibinfo{person}{Bin Cheng},
  \bibinfo{person}{Jiashu Xing}, \bibinfo{person}{Yangtao Wang},
  \bibinfo{person}{Tianheng Cheng}, \bibinfo{person}{Li Liu},
  \bibinfo{person}{Minwei Ran}, {and} \bibinfo{person}{Zekang Li}.}
  \bibinfo{year}{2019}\natexlab{}.
\newblock \showarticletitle{An End-to-End Automatic Cloud Database Tuning
  System Using Deep Reinforcement Learning}. In
  \bibinfo{booktitle}{\emph{Proceedings of the 2019 International Conference on
  Management of Data}} (Amsterdam, Netherlands) \emph{(\bibinfo{series}{SIGMOD
  '19})}. \bibinfo{publisher}{Association for Computing Machinery},
  \bibinfo{address}{New York, NY, USA}, \bibinfo{pages}{415–432}.
\newblock
\showISBNx{9781450356435}
\urldef\tempurl%
\url{https://doi.org/10.1145/3299869.3300085}
\showDOI{\tempurl}


\bibitem[Zhang et~al\mbox{.}(2022a)]%
        {Bin2022}
\bibfield{author}{\bibinfo{person}{Xinyi Zhang}, \bibinfo{person}{Zhuo Chang},
  \bibinfo{person}{Yang Li}, \bibinfo{person}{Hong Wu}, \bibinfo{person}{Jian
  Tan}, \bibinfo{person}{Feifei Li}, {and} \bibinfo{person}{Bin Cui}.}
  \bibinfo{year}{2022}\natexlab{a}.
\newblock \showarticletitle{Facilitating database tuning with hyper-parameter
  optimization: a comprehensive experimental evaluation}.
\newblock \bibinfo{journal}{\emph{Proc. VLDB Endow.}} \bibinfo{volume}{15},
  \bibinfo{number}{9} (\bibinfo{date}{may} \bibinfo{year}{2022}),
  \bibinfo{pages}{1808--1821}.
\newblock
\showISSN{2150-8097}
\urldef\tempurl%
\url{https://doi.org/10.14778/3538598.3538604}
\showDOI{\tempurl}


\bibitem[Zhang et~al\mbox{.}(2021)]%
        {zhang2021restune}
\bibfield{author}{\bibinfo{person}{Xinyi Zhang}, \bibinfo{person}{Hong Wu},
  \bibinfo{person}{Zhuo Chang}, \bibinfo{person}{Shuowei Jin},
  \bibinfo{person}{Jian Tan}, \bibinfo{person}{Feifei Li},
  \bibinfo{person}{Tieying Zhang}, {and} \bibinfo{person}{Bin Cui}.}
  \bibinfo{year}{2021}\natexlab{}.
\newblock \showarticletitle{Restune: Resource oriented tuning boosted by
  meta-learning for cloud databases}. In \bibinfo{booktitle}{\emph{Proceedings
  of the 2021 international conference on management of data}}.
  \bibinfo{pages}{2102--2114}.
\newblock


\bibitem[Zhang et~al\mbox{.}(2022b)]%
        {zhang2022towards}
\bibfield{author}{\bibinfo{person}{Xinyi Zhang}, \bibinfo{person}{Hong Wu},
  \bibinfo{person}{Yang Li}, \bibinfo{person}{Jian Tan},
  \bibinfo{person}{Feifei Li}, {and} \bibinfo{person}{Bin Cui}.}
  \bibinfo{year}{2022}\natexlab{b}.
\newblock \showarticletitle{Towards dynamic and safe configuration tuning for
  cloud databases}. In \bibinfo{booktitle}{\emph{Proceedings of the 2022
  International Conference on Management of Data}}. \bibinfo{pages}{631--645}.
\newblock


\bibitem[Zhao et~al\mbox{.}(2023)]%
        {zhao2023automatic}
\bibfield{author}{\bibinfo{person}{Xinyang Zhao}, \bibinfo{person}{Xuanhe
  Zhou}, {and} \bibinfo{person}{Guoliang Li}.} \bibinfo{year}{2023}\natexlab{}.
\newblock \showarticletitle{Automatic database knob tuning: a survey}.
\newblock \bibinfo{journal}{\emph{IEEE Transactions on Knowledge and Data
  Engineering}} \bibinfo{volume}{35}, \bibinfo{number}{12}
  (\bibinfo{year}{2023}), \bibinfo{pages}{12470--12490}.
\newblock


\bibitem[Zhou(2012)]%
        {zhou2012ensemble}
\bibfield{author}{\bibinfo{person}{Zhi-Hua Zhou}.}
  \bibinfo{year}{2012}\natexlab{}.
\newblock \bibinfo{booktitle}{\emph{Ensemble methods: foundations and
  algorithms}}.
\newblock \bibinfo{publisher}{CRC press}.
\newblock


\bibitem[Zhu et~al\mbox{.}(2024)]%
        {pilotscope24}
\bibfield{author}{\bibinfo{person}{Rong Zhu}, \bibinfo{person}{Lianggui Weng},
  \bibinfo{person}{Wenqing Wei}, \bibinfo{person}{Di Wu},
  \bibinfo{person}{Jiazhen Peng}, \bibinfo{person}{Yifan Wang},
  \bibinfo{person}{Bolin Ding}, \bibinfo{person}{Defu Lian},
  \bibinfo{person}{Bolong Zheng}, {and} \bibinfo{person}{Jingren Zhou}.}
  \bibinfo{year}{2024}\natexlab{}.
\newblock \showarticletitle{PilotScope: Steering Databases with Machine
  Learning Drivers}.
\newblock \bibinfo{journal}{\emph{Proc. VLDB Endow.}} \bibinfo{volume}{17},
  \bibinfo{number}{5} (\bibinfo{date}{may} \bibinfo{year}{2024}),
  \bibinfo{pages}{980--993}.
\newblock
\showISSN{2150-8097}
\urldef\tempurl%
\url{https://doi.org/10.14778/3641204.3641209}
\showDOI{\tempurl}


\end{thebibliography}

\end{document}